\documentclass[lettersize,journal]{IEEEtran}
\usepackage{amsmath,amsfonts,amssymb}
\usepackage{algorithm}
\usepackage{algpseudocode}
\usepackage{array}
\usepackage[caption=false,font=normalsize,labelfont=sf,textfont=sf]{subfig}
\usepackage{textcomp}
\usepackage{stfloats}
\usepackage{url}
\usepackage{verbatim}
\usepackage{graphicx}
\usepackage{xcolor}
\usepackage{mathtools}
\usepackage{booktabs}
\usepackage{multirow}
\usepackage{placeins}
\usepackage{hyperref}
\graphicspath{{./}{picture/}{application/}}

\usepackage{makecell}
\usepackage{amsthm}
\usepackage[numbers,sort&compress]{natbib}

\usepackage[table]{xcolor}   
\usepackage{amssymb}         

\definecolor{okblue}{RGB}{0,51,153}    
\definecolor{nored}{RGB}{153,0,0}      
\newcommand{\cmark}{\textcolor{okblue}{\checkmark}}
\newcommand{\xmark}{\textcolor{nored}{$\times$}}

\newtheorem{theorem}{Theorem}[section]
\newtheorem{proposition}[theorem]{Proposition}
\newtheorem{lemma}[theorem]{Lemma}

\newtheorem{assumption}[theorem]{Assumption}
\newtheorem{remark}[theorem]{Remark}

\begin{document}

\title{SCAN: Neural Additive Networks for Scalable High-Order Component Selection and Estimation}

\author{Seokhun~Park, Kunwoong~Kim, and Yongdai~Kim
\thanks{This work has been submitted to the IEEE for possible publication.
Copyright may be transferred without notice, after which this version may
no longer be accessible.}
\thanks{S. Park and Y. Kim are with the Department of Statistics, Seoul National University, Seoul 08826, Republic of Korea.}
\thanks{K. Kim is with KAIST AI, Seoul 06764, Republic of Korea.}}

\markboth{Preprint. Under review.}%
{Park \MakeLowercase{\textit{et al.}}: SCAN}

\maketitle

\begin{abstract}
Functional ANOVA models express a prediction as a sum of main effects and interactions, but existing neural functional ANOVA models require the interactions to be specified in advance, which forces a trade-off between missing higher-order structure and the cost of enumerating many interactions.
We propose SCAN (Scalable Component Additive Networks), which learns which features each basis function combines, so that components of different orders are selected and estimated jointly without enumerating candidates.
The learned component index sets form a small candidate collection on which post-hoc identification is carried out efficiently.
A global component-selection parameter shares information across basis functions, and truncating negligible component-selection weights gives a sparse model that is identified after training.
We establish universal approximation, a minimax optimal rate up to a logarithmic factor, and component selection consistency, which together imply that the discovered components contain all identified true components with probability tending to one.
In synthetic experiments, SCAN learns components whose maximum order is close to the true one, and recovers higher-order components more accurately than the baseline models, with post-hoc identification taking a few seconds for 100 inputs.
SCAN also outperforms most neural functional ANOVA models in prediction on the benchmark datasets and trains at a much lower cost than exhaustive second-order models.
We also apply SCAN to predict treatment response in breast cancer from 12,502 gene-expression variables.
\end{abstract}

\begin{IEEEkeywords}
Component selection, functional ANOVA models, interpretable machine learning, neural networks
\end{IEEEkeywords}

\section{Introduction}

Understanding a prediction model often requires more than identifying important input features.
In particular, when there are interactions between input features, the relation between features and output cannot be understood from the importance of each input feature.
Functional ANOVA model (\cite{hooker_diag, wahba1995smoothing}) decomposes a high-dimensional function as a sum of functions called components, each depending on a subset of the inputs, and so important interactions and their relations to the output can be identified. 

Neural functional ANOVA models, including Neural Additive Models (NAM, \cite{agarwal2021neural}), Neural Basis Models (NBM, \cite{radenovic2022neural}), and ANOVA-TPNN (\cite{park2025tensor}), use neural networks to estimate the component functions.
Their computational cost, however, becomes large rapidly as the number of input features increases.
For example, a prediction model with a 700-dimensional input feature has roughly 245,000 possible second-order interactions, and exponentially many more components are added for higher-order interactions.
Thus, the existing neural functional ANOVA models focus only on lower-order interactions such as the main effects and the second order interactions. 

Component selection offers a way to avoid considering exponentially many components.
NODE-GAM (\cite{chang2021node}) can select features in each component, but the maximum number of features in each component (i.e. the maximum order of interactions) is fixed at 1 or 2 because it uses tensor-product basis functions which become computationally unstable easily for high-order interactions.
Recently, \cite{park2025bayesian} develops a Bayesian neural functional ANOVA model to explore higher-order interactions.
The MCMC algorithm, however, is not easily scaled up since storing MCMC samples is expensive and mini-batch learning is not available. 
\begin{table}[t]
\centering
\caption{\footnotesize \textbf{Comparison of functional ANOVA models.} Theory: convergence rate and component selection consistency.}
\label{tab:model_comparison}
\setlength{\tabcolsep}{3pt}   
\resizebox{\columnwidth}{!}{%
\begin{tabular}{lccccc}
\toprule
Model & \makecell{Max.\\order} & \makecell{No\\enum.} & \makecell{Mini-\\batch} & Selection & Theory \\
\midrule
NAM (\cite{agarwal2021neural})      & Predefined & \xmark & \cmark & \xmark & \xmark \\
NBM (\cite{radenovic2022neural})    & Predefined & \xmark & \cmark & \xmark & \xmark \\
ANOVA-TPNN (\cite{park2025tensor})  & Predefined & \xmark & \cmark & \xmark & \xmark \\
EBM (\cite{nori2019interpretml})   & Predefined & \xmark & \xmark & \cmark & \xmark \\
NODE-GAM (\cite{chang2021node})   & Predefined & \cmark & \cmark & \cmark & \xmark \\
\midrule
SCAN (Ours) & \makecell{Data-adaptive} & \cmark & \cmark & \cmark & \cmark \\
\bottomrule
\end{tabular}%
}
\end{table}

The aim of this paper is to propose a novel neural functional ANOVA model, called \textbf{SCAN (Scalable Component Additive Networks)},
which can be trained by a standard mini-batch gradient descent algorithm and can identify higher-order interactions with manageable computational and memory costs.
For this purpose, we develop a special basis function that depends on a linear combination of univariate transformations of the input features, where the weights in the linear combination are learnable and sparse.
Thus, each basis function uses only a small number of input features, which provides data-adaptive interaction selection as a by-product.
Table~\ref{tab:model_comparison} compares SCAN with existing functional ANOVA models in terms of whether the maximum order of components must be predefined, whether all candidate components are enumerated, whether mini-batch training is possible, whether components are selected data-adaptively, and whether theoretical guarantees are available.
Existing models require the maximum order of components to be predefined, typically at one or two, and either enumerate all candidate components or lack theoretical guarantees.
In contrast, SCAN determines the orders of components data-adaptively by mini-batch training without enumerating candidates, and comes with theoretical guarantees for estimation and component selection.

SCAN assumes that the prediction model is a sum of these basis functions, trains the parameters in the prediction model by a mini-batch gradient descent algorithm and reorganizes the trained basis function into the functional ANOVA model.
Note that SCAN can identify important components data-adaptively because the basis functions do so.
Theoretically, we prove statistical optimality as well as theoretical guarantees for component selection. 
The main contributions are as follows.
\begin{itemize}
    \item We propose a neural functional ANOVA architecture that can be trained by a standard mini-batch gradient-based method
    but can jointly select and estimate components without enumerating a prespecified set of interactions. 
    
    \item We establish the universal approximation for H\"older continuous functions and a minimax optimal convergence rate up to a logarithmic factor for the least squares estimator.
    Under additional assumptions, we also show the component selection consistency.

    \item We evaluate our method in view of prediction, component selection, computational cost, and interpretation on multiple benchmark datasets and a high-dimensional gene-expression application.
    The empirical results show improved performance of SCAN compared to the existing neural functional ANOVA models and illustrate how SCAN learns an interpretable representation from high-dimensional input features.
\end{itemize}

\section{Background and Preliminaries}

\subsection{Notations}\label{sec:notation}
Throughout the paper, $\mathbf{x} = (x_1, \ldots, x_p)^\top$ denotes an input feature vector taking values in $\mathcal{X} =\mathcal{X}_{1}\times \cdots \times \mathcal{X}_{p} \subseteq [-a, a]^p$ for some constant $a > 0$.
The index set of the input features is written as $[p] = \{1, \ldots, p\}$.
Given a subset $S \subseteq [p]$, $\mathbf{x}_S=(x_j, j\in S)$ denotes the subvector of $\mathbf{x}$ formed by the coordinates in $S$, and $f_S(\mathbf{x}_S)$ denotes a function whose value depends on $\mathbf{x}$ only through $\mathbf{x}_S$.
Let $\mathbf{x}_{i} = (x_{i,1}, \ldots, x_{i,p})^{\top}$, $i = 1, \ldots, n$, be observed input feature vectors.
For $1 \leq r < \infty$, the empirical $r$ norm of a function $f$ is defined by
$\Vert f \Vert_{r,n} := \big( n^{-1} \sum_{i=1}^{n} |f(\mathbf{x}_{i})|^{r} \big)^{1/r}$,
and its population counterpart with respect to a probability measure $\mu$ on $\mathcal{X}$ is
$\Vert f \Vert_{r,\mu} := \big( \int_{\mathcal{X}} |f(\mathbf{x})|^{r} \, \mu(d\mathbf{x}) \big)^{1/r}$.
For a vector $\mathbf{e} = (e_1, \ldots, e_n)^\top$, we write $\Vert \mathbf{e} \Vert_{1} := \sum_{i=1}^{n} |e_{i}|$ for its $\ell_1$ norm and $\Vert \mathbf{e} \Vert_{2} := \big( \sum_{i=1}^{n} e_{i}^{2} \big)^{1/2}$ for its $\ell_2$ norm.
For a set $A$, let $\mathcal{P}(A)$ denote the power set of $A$, i.e., the collection of all subsets of $A$.
Let $\mu_j$ denote the marginal distribution defined on $\mathcal{X}_{j}$ and let $\bar{\mu} := \mu_1 \otimes \cdots \otimes \mu_p$ be the product of the marginals.

\subsection{Functional ANOVA Model}

The functional ANOVA model decomposes a high-dimensional function into a sum of low-dimensional functions (components):
\begin{align}
f(\mathbf{x}) = \beta_0 + \sum_j f_j(x_j) + \sum_{j<k} f_{jk}(x_j, x_k) + \cdots,
\end{align}
where $f_j$ are the main effects, $f_{jk}$ are the second-order interactions and so on.
A generalized additive model (GAM) (\cite{hastie2017generalized}) is the special case containing only the main effects.

Traditional approaches for learning functional ANOVA models include smoothing splines (SSANOVA, \cite{wahba1995smoothing}), component-wise penalties (\cite{cosso}), and basis expansion methods (\cite{rugamer2024scalable}).
These methods work well for data of moderate size but become computationally prohibitive for high-dimensional/large sized data. 

To improve scalability, neural network-based functional ANOVA models have recently been actively studied, which model each component by a specially designed neural network and estimate all components functions simultaneously by a gradient-based optimization.
Representative examples are NAM (\cite{agarwal2021neural}), NBM (\cite{radenovic2022neural}), and ANOVA-TPNN (\cite{park2025tensor}).

These examples (\cite{agarwal2021neural,radenovic2022neural,park2025tensor}), however, require the set of candidate components to be specified before training.
When the dimension of input feature is large and/or the maximum order of interactions for candidate components is large, the number of candidate components to be estimated becomes prohibitively large. 
Thus, they fix the maximum order of components at 1 or 2, and thus
are not able to identify important higher-order interactions.

To overcome this limitation, several models have been proposed that 
select and estimate signal components.
NODE-GAM (\cite{chang2021node}) builds on Neural Oblivious Decision Ensembles (NODE, \cite{popov2019neural}), an ensemble of neural oblivious decision trees (NODTs), where features used in each NODT of the ensemble are selected data-adaptively.
NODE-GAM restricts each NODT to use at most two features because of computational issues.
More recently, \cite{kishimoto2024neural} introduces a similar feature-selection approach into NAM and NBM.
However, both approaches still require the maximum order of the components to be fixed in advance at 1 or 2 because of a similar reason to NODE-GAM.

Bayesian-TPNN (\cite{park2025bayesian}) treats the network architecture of each basis function as a learnable parameter and explores the relevant components via an MCMC algorithm, without enumerating all of the candidate components.
However, this approach is not easily scalable.
First, it is computationally demanding to make an MCMC algorithm to
 explore adequately the space of all candidate components that is exponentially growing with respect to the input feature dimension.
Second, mini-batch computation is not available for the MCMC update, which limits the sample sizes that can be handled within a given computational budget.

\begin{figure*}[t]
    \centering
    \includegraphics[width=0.7\linewidth]{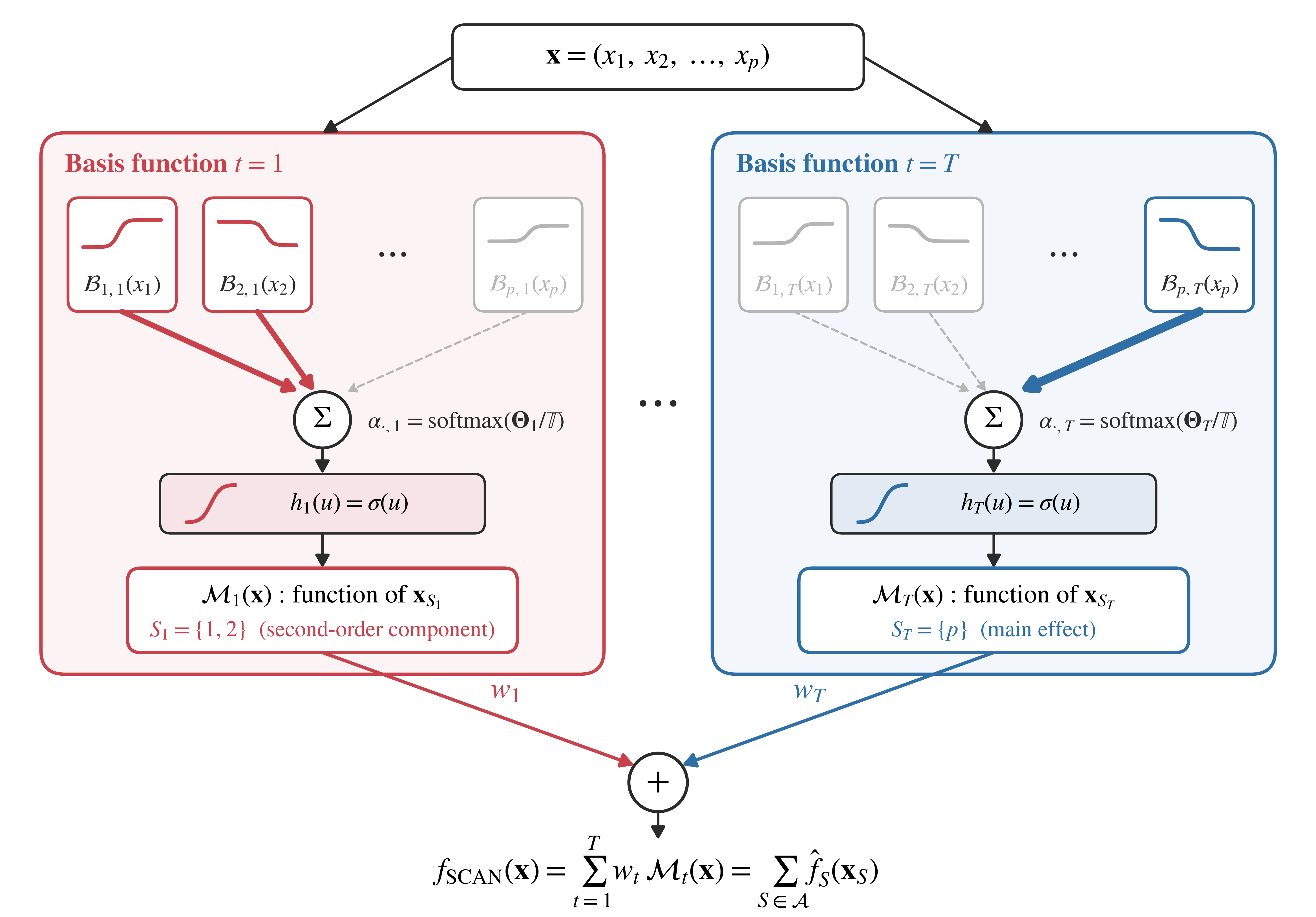}
    \caption{\small Architecture of SCAN. Each basis function combines univariate unit functions through learnable component-selection weights.
    The selected component index sets determine the component structure, and a weighted sum of the basis functions gives the prediction.}
    \label{fig:architecture}
\end{figure*}

\section{SCAN: Scalable Component Additive Networks}

\subsection{Main Idea and Architecture}

SCAN learns which features each basis function uses, and thereby is able to select signal components as a by-product.
Hence, SCAN does not require a predefined set of candidate interactions (e.g. a maximum order of interactions).
SCAN lets the prediction model have the following form:
\begin{align}
f_{\text{SCAN}}(\mathbf{x}) = \sum_{t=1}^{T} w_{t}\mathcal{M}_t(\mathbf{x}),
\end{align}
where each basis function $\mathcal{M}_{t}$ is further modeled as
\begin{align}
\mathcal{M}_t(\mathbf{x}) = h\left(\sum_{j=1}^p \alpha_{j,t}\mathcal{B}_{j,t}(x_j)\right).
\end{align}
Here, $\mathcal{B}_{j,t}(\cdot)$ is a univariate unit function and $h(\cdot)$ is the sigmoid function.
The component-selection weights $\alpha_{j,t}$ determine the contribution of feature $j$ to $\mathcal{M}_t$.
If we make $\alpha_{j,t}$ sparse, only a small portion of input features are involved in
$\mathcal{M}_t$ and so the component selection is achieved.

\subsection{Univariate Unit Functions}

For $\mathcal{B}_{j,t}$, we use a depth-1 smooth decision tree,
\begin{align*}
\small
\mathcal{B}_{j,t}(x_j) := \beta_{j,t,1}\left(1 - \sigma\left(\frac{x_j - b_{j,t}}{\gamma_{j,t}}\right)\right) + \beta_{j,t,2}\sigma\left(\frac{x_j - b_{j,t}}{\gamma_{j,t}}\right)
\end{align*}
where $\sigma(u) := 1/(1 + e^{-u})$ is the sigmoid function, and $(\beta_{j,t,1}, \beta_{j,t,2}, b_{j,t}, \gamma_{j,t})$ are learnable parameters.
This form is compact, requiring only four parameters per unit, and its output is bounded between the two leaf values, which prevents unstable extrapolation.
The comparison in Section~\ref{app:choice_basis} shows that this choice performs favorably among the univariate unit functions evaluated.

\subsection{Component-selection Weights}

\begin{figure}
    \centering
    \includegraphics[width=1.0\linewidth]{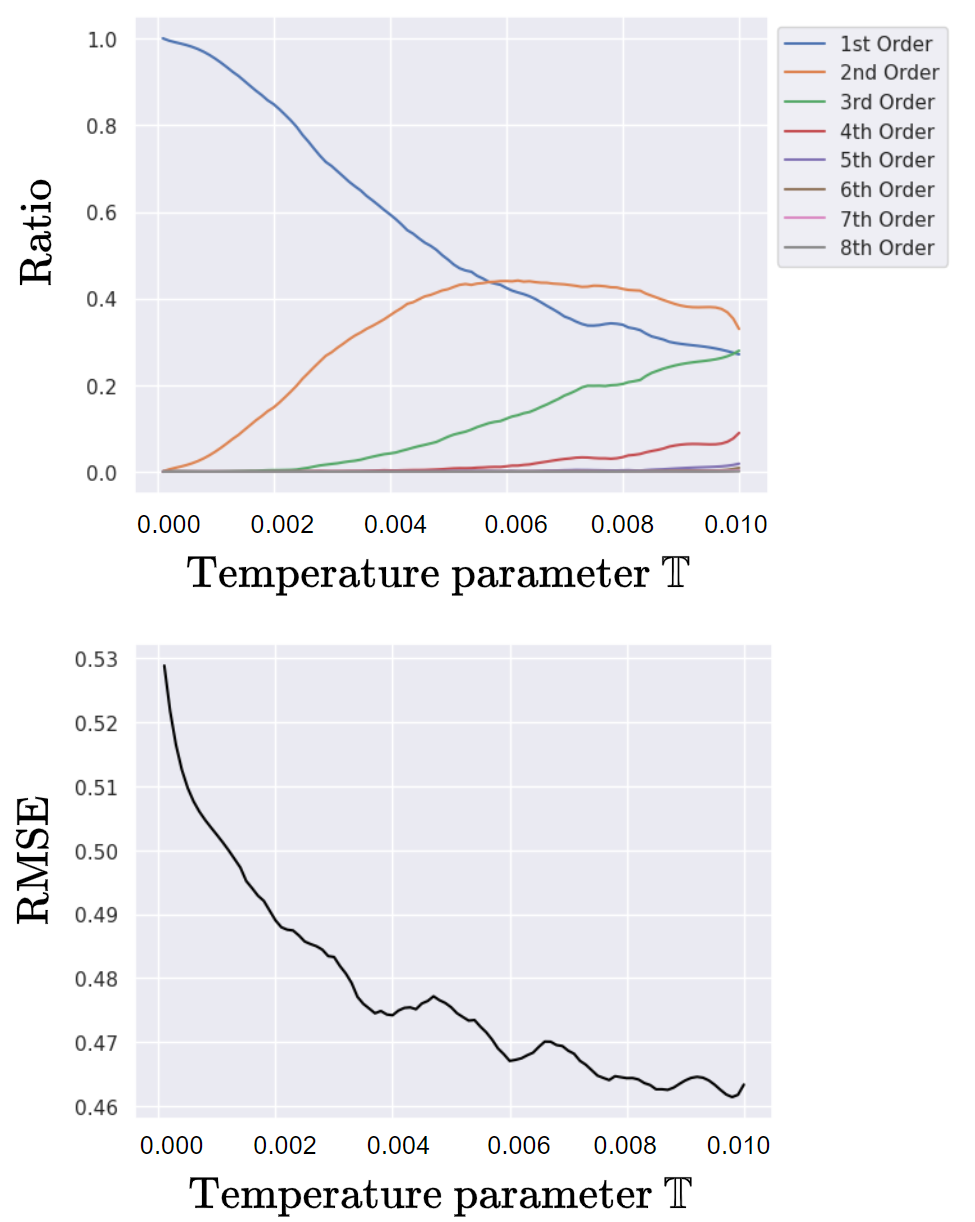}
    \caption{\textbf{Effect of the temperature $\mathbb{T}$ on the orders of the estimated components on \textsc{Calhousing} dataset.} The top panel shows the proportion of the basis functions estimating components of each order, and the bottom panel shows the prediction performance (RMSE) of SCAN as $\mathbb{T}$ varies.}
    \label{fig:T_example}
\end{figure}

The component-selection weights $\alpha_{j,t}$s must be nonnegative and their scales should be controlled to avoid the identifiability issue with $\mathcal{B}_{j,t}.$ So we let $\alpha_t=(\alpha_{j,t}, j\in [p])$ lie on the $p$-dimensional simplex.
In addition, we want to make $\alpha_t$ be sparse. One may put the $\ell_0$ penalty on $\alpha_t$ 
(i.e. $\sum_{j=1}^p \mathbb{I}(\alpha_{j,t}\ne 0)$),
but a gradient-based learning would not be feasible. 
Motivated by \cite{chang2021node}, instead of the $\ell_0$ penalty, we use a smooth surrogate of the $\ell_0$ penalized $\alpha_t$ on the simplex as:
\begin{align}
\alpha_t = \text{softmax}\left(\frac{\boldsymbol{\Theta}_t}{\mathbb{T}}\right),
\end{align}
where $\boldsymbol{\Theta}_{t} := (\theta_{1,t},...,\theta_{p,t})^{\top} \in \mathbb{R}^p$ is a learnable parameter vector and $\mathbb{T} > 0$ is a temperature hyperparameter. 
Note that the temperature $\mathbb{T}$ controls the sparsity of the component-selection weights.
A smaller temperature favors a few dominant component-selection weights and hence yields a higher sparsity after truncating small component-selection weights, whereas a larger temperature results in less sparsity. 
By controlling the temperature $\mathbb{T}$ appropriately, we can delete unnecessary higher-order interactions 
and so identifying and estimating higher-order signal components becomes possible with a reasonable amount of computing resources. Figure~\ref{fig:T_example} illustrates the resulting relationship between component order and prediction performance.

\begin{remark}[Motivation of SCAN]
The architecture of SCAN is motivated by Kolmogorov-Arnold Networks (KAN, \cite{liu2024kan}),
which are based on the Kolmogorov-Arnold representation theorem (\cite{kolmogorov1961representation})
and represent a multivariate function as sums and compositions of learnable univariate functions,
i.e., 
$$
f(\mathbf{x}) = \sum_{q} \Phi_q \big( \sum_{j=1}^{p} \phi_{q,j}(x_j) \big),
$$
where $\phi_{q,j}:\mathbb{R}\rightarrow\mathbb{R}$ are univariate inner functions applied to each input variable $x_j$,
and $\Phi_q:\mathbb{R}\rightarrow\mathbb{R}$ are univariate outer functions applied to the aggregated inner representations.
SCAN adopts this two-level structure, where the inner and outer univariate functions correspond to
$\alpha_{j,t}\mathcal{B}_{j,t}$ and $w_t h$, respectively.
This compositional structure alone does not specify a sparse functional ANOVA decomposition.
SCAN introduces the explicit component-selection weights $\alpha_t$, so that each retained basis function depends only on a small number of selected features provided that $\alpha_t$ is sparse. 
Finally, the resulting prediction model is rewritten as the functional ANOVA model where only a small number of components are involved. 
\end{remark}

\subsection{Reformulation to the functional ANOVA model}

The key to the automatic component selection of SCAN lies in the sparsity of the learned component-selection weights $\alpha_{j,t}$.
During training, the component-selection weights $(\alpha_{1,t}, \ldots, \alpha_{p,t})$ with a small value of the temperature $\mathbb{T}$ tend to concentrate on a few features, while the others approach zero.
However, since the softmax function never produces exact zeros, each $\mathcal{M}_t$ still depends on all features in a strict sense.
To obtain a meaningful functional ANOVA representation, we truncate negligible component-selection weights at inference.
That is, for a small threshold $\epsilon_{\alpha} > 0$, we use
\begin{align}
\tilde{\alpha}_{j,t} := \alpha_{j,t}\, \mathbb{I}(\alpha_{j,t} > \epsilon_{\alpha})
\end{align}
in place of $\alpha_{j,t}$ after training.
Consequently, each basis function $\mathcal{M}_t$ depends only on the subset $\mathbf{x}_{S_t}$ of features, where
\begin{align}
S_t = \{j \in [p] : \alpha_{j,t} > \epsilon_{\alpha}\}.
\end{align}
Truncation makes the feature set $S_t$ of $\mathcal{M}_t$ sparse,
but the resulting predictions may differ from the untruncated ones.
However, Theorem \ref{thm:conv_rate} proves that the truncation does not hamper the theoretical optimality of SCAN.
More discussion for the choice of $\epsilon_{\alpha}$ is given in Remark~\ref{rem:eps_choice}.

After the truncation, the $T$ basis functions have their own feature sets.
By collecting basis functions having the same feature set, we can construct the estimate of the corresponding component. 
That is, the trained SCAN model can be rewritten as the following functional ANOVA model:
\begin{align}
\hat{f}_{\text{SCAN}}(\mathbf{x} \mid \boldsymbol{\psi},\epsilon_{\alpha}) = \sum_{S \in \mathcal{A}} \hat{f}_{\text{SCAN},S}(\mathbf{x}_S),
\label{eq:scan}
\end{align}
where $\mathcal{A} = \{S_1, \ldots, S_T\}$ is the set of the feature sets, and the estimated component for $S \subseteq [p]$ is given by
\begin{align}
\hat{f}_{\text{SCAN},S}(\mathbf{x}_S) = \sum_{t \in [T] : S_t = S} w_t \mathcal{M}_t(\mathbf{x} \mid \boldsymbol{\zeta}_t).
\label{eq:comp_unid}
\end{align}
Here, $\boldsymbol{\psi} := \{w_t, \boldsymbol{\zeta}_t : t \in [T]\}$ denotes all learnable parameters, where $\boldsymbol{\zeta}_t := \{\boldsymbol{\Theta}_t, \mathbf{b}_t\}$ denotes the parameters in the basis function $\mathcal{M}_t$ with $\mathbf{b}_t := \{\beta_{j,t,1}, \beta_{j,t,2}, b_{j,t}, \gamma_{j,t}\}_{j=1}^{p}$.

\subsection{Global component-selection parameter}
\label{sec:global_prior}

By equation (\ref{eq:comp_unid}), each component $\hat f_{\text{SCAN},S}$ is estimated by the sum of the basis functions whose feature sets are equal to $S$.
Hence, accurately estimating a component requires multiple basis functions having the same feature set.
However, the number of parameters in the component selection weights is proportional to the number of the basis functions $T$,
and thus their estimates would become unstable when $T$ is large which results
in selection of spurious components.
This problem becomes more apparent when most of the features are irrelevant, since irrelevant features selected by chance accumulate across the basis functions and lead to overfitting.

To stabilize the component-selection weights and 
to encourage the basis functions to select their feature sets from a common set of relevant features, 
we introduce a global component-selection parameter $\mathbf{g} := (g_1, \ldots, g_p)^\top \in \mathbb{R}^{p}$, a learnable parameter vector shared by all basis functions, and define the component-selection weights as
\begin{align}
(\alpha_{1,t}, \ldots, \alpha_{p,t})^\top = \text{softmax}\left( \frac{\boldsymbol{\Theta}_t}{\mathbb{T}} + \frac{\mathbf{g}}{\mathbb{T}_g} \right),
\label{eq:dual_temp}
\end{align}
where $\mathbb{T}_g > 0$ is the temperature for the global component-selection parameter.
Since $\mathbf{g}$ is shared by all basis functions, its gradient is the sum of the gradients from all $T$ basis functions.
This aggregation allows evidence for the feature sets to be shared across the basis functions and so makes the selection of feature sets more stable.
The global temperature $\mathbb{T}_g$ plays a role of eliminating irrelevant features from
all basis functions while the local temperature $\mathbb{T}$ influences the selection of
feature sets in each basis.
This motivates the joint use of the local temperature $\mathbb{T}$ and global temperature $\mathbb{T}_{g}$, whose roles are examined in Section~\ref{sec:exp_madelon}.

The two temperatures therefore play complementary roles.
The local temperature $\mathbb{T}$ controls concentration within a basis function, whereas the global temperature $\mathbb{T}_g$ controls the influence of the global component-selection parameter across basis functions.
In practice, we anneal both temperatures from large values to their final values during training, which we found to be crucial for stable component selection.
The details of the annealing schedule are provided in Section \ref{app:annealing}.

\subsection{Unique Identification of Components}
\label{sec:identification}

A remaining problem is that the functional ANOVA decomposition in (\ref{eq:scan}) is not unique.
The following is an example of nonuniqueness of the functional ANOVA decomposition:
\begin{align*}
f(x_1, x_2) &= x_1 + x_2 + x_1 x_2 \\
&= -x_1 + x_2 + x_1 (x_2 + 2).
\end{align*}
Consequently, the estimated components $\hat f_S$ in (\ref{eq:scan})
cannot be used directly for interpretation. 

To resolve this issue, we apply a post-hoc unique identification (\cite{lengerich2020purifying, chang2021node}) of the components to the estimated SCAN predictor.
Let $\nu = \nu_1 \times \cdots \times \nu_p$, where $\nu_j$ is the empirical distribution 
on $\{x_{1,j},\ldots,x_{n,j}\}.$
For a function $g$ on $\mathcal{X}$ and $U \subseteq [p]$, let 
$$(\mathbb{E}_{\nu} g)(\mathbf{x}_U) := \int g(\mathbf{x}_U, \mathbf{z}_{-U})\, \nu_{-U}(d\mathbf{z}_{-U})$$ with $\nu_{-U} := \prod_{j \notin U} \nu_j.$
If we require that each $\hat{f}_{\text{SCAN},S}$ satisfies  
$(\mathbb{E}_{\nu} \hat{f}_{\text{SCAN},S})(\mathbf{x}_U)=0$ for all $\mathbf{x}_U$ and all $U\subset S$ with $|S-U|=1,$
then $\hat{f}_{\text{SCAN},S}$s are uniquely defined (see Proposition 3.1 in \cite{park2025tensor}).

Now, we explain how we transform the estimated components $\hat{f}_{\text{SCAN},S}, S\in \mathcal{A}$ into those satisfying the above uniqueness condition. 
First, for a given function $g$
on $\mathcal{X},$ we define the identifiable projection of $g$ on the feature set $S\subset [p]$ as
\begin{align}
\Pi_S g(\mathbf{x}) := \sum_{U \subseteq S} (-1)^{|S| - |U|} (\mathbb{E}_{\nu} g)(\mathbf{x}_U).
\label{eq:purify}
\end{align}
Proposition \ref{prop:projection-0} in Section \ref{app:projection} shows that $\Pi_S \mathcal{M}_t =0 $ when $S \not \subseteq S_{t}.$
Let $\mathcal{D}(\mathcal{A})$ denote the downward closure of $\mathcal{A}$, i.e.,
\begin{align*}
\mathcal{D}(\mathcal{A}) := \{ S \subseteq [p] : S \subseteq S' \text{ for some } S' \in \mathcal{A} \},
\end{align*}
which is the collection of all subsets (including $\emptyset$) of the discovered feature sets.
Since $\hat{f}_{\text{SCAN}} = \sum_{t=1}^{T} w_t \mathcal{M}_t$ and $\Pi_S \mathcal{M}_t = 0$ whenever $S \not\subseteq S_t$, we have $\Pi_S \hat{f}_{\text{SCAN}}(\mathbf{x}) = 0$ for every $S \notin \mathcal{D}(\mathcal{A})$ and so 
\begin{align}
\hat{f}_{\text{SCAN}}(\mathbf{x})
= \sum_{S \subseteq [p]} \Pi_S \hat{f}_{\text{SCAN}}(\mathbf{x})
= \sum_{S \in \mathcal{D}(\mathcal{A})} \Pi_S \hat{f}_{\text{SCAN}}(\mathbf{x}).
\label{eq:closure}
\end{align}

Moreover, each $\Pi_S \hat{f}_{\text{SCAN}}$ with $S \in \mathcal{D}(\mathcal{A}) \setminus \{\emptyset\}$ satisfies the identifiability condition, that is,
\begin{align*}
\int_{\mathcal{X}_{j}} \Pi_S \hat{f}_{\text{SCAN}}(\mathbf{x}_S)\, \nu_j(dx_j) = 0
\qquad \text{for every } j\in S,
\end{align*}
which is proved in Section \ref{app:projection}.
For each $S \subseteq [p]$, we call $\hat f^{*}_S := \Pi_S \hat{f}_{\text{SCAN}}$ the identified component of SCAN
for the feature set $S.$ Note that $\hat f^{*}_S$
is an estimate of the true identified component $f_{0,S}^{*}$ defined in Section \ref{sec:convergence}. \\
$\newline$
\textbf{Computation.}
The post-hoc identification uses the selected feature sets rather than all subsets of the $p$ inputs.
Its cost therefore depends on the number and sizes of the discovered feature sets (see Section~\ref{app:cost_identification}).
Since $\Pi_S$ is linear and $\Pi_S \hat f_{\text{SCAN},S'} = 0$ unless $S \subseteq S'$, we have
\begin{align*}
\hat f^{*}_S := \Pi_{S}\hat{f}_{\text{SCAN}} = \sum_{S' \in \mathcal{A},\, S' \supseteq S} \Pi_S \hat f_{\text{SCAN},S'},
\end{align*}
where $\hat{f}_{\text{SCAN},S'}$s are defined in (\ref{eq:comp_unid}), so that only the subsets of the discovered feature sets in $\mathcal{A}$ need to be considered.
Moreover, each $\Pi_S \hat f_{\text{SCAN},S'}$ involves only the coordinates in $S'$.
In practice, we compute the identified components as in Algorithm \ref{alg:purify}.
\begin{algorithm}[t]
\caption{Post-hoc identification of SCAN}
\label{alg:purify}
\begin{algorithmic}[1]
\Require Discovered components $\mathcal{A}$, raw components
  $\{\hat f_{\text{SCAN},S'}\}_{S' \in \mathcal{A}}$, observed data
  $\{\mathbf{x}_i\}_{i=1}^n$, number of points $m$
\Ensure Identified components $\{\hat f^{*}_S\}_{S \in \mathcal{D}(\mathcal{A})}$
\State Sample $\mathbf{z}^{(1)}, \ldots, \mathbf{z}^{(m)}$, where each
  coordinate $z^{(k)}_j$ is sampled from $\{x_{1,j}, \ldots, x_{n,j}\}$
\State Initialize $\hat f^{*}_S \gets 0$ for all $S \in \mathcal{D}(\mathcal{A})$
\For{$S' \in \mathcal{A}$}
  \For{$U \subseteq S'$} \Comment{marginalize}
    \State ${(\hat{\mathbb{E}}_{\nu} \hat f_{\text{SCAN},S'})(\mathbf{x}_U) \gets
       \frac{1}{m} \sum_{k=1}^{m} \hat f_{\text{SCAN},S'}\big(\mathbf{x}_U, \mathbf{z}^{(k)}_{S' \setminus U}\big)}$
  \EndFor
  \For{$S \subseteq S'$} \Comment{inclusion-exclusion (\ref{eq:purify})}
    \State $\Pi_S \hat f_{\text{SCAN},S'} \gets \sum_{U \subseteq S} (-1)^{|S|-|U|}
      (\hat{\mathbb{E}}_{\nu} \hat f_{\text{SCAN},S'})(\mathbf{x}_U)$
    \State $\hat f^{*}_S \gets \hat f^{*}_S + \Pi_S \hat f_{\text{SCAN},S'}$
      \Comment{aggregate}
  \EndFor
\EndFor
\State \Return $\{\hat f^{*}_S\}_{S \in \mathcal{D}(\mathcal{A})}$
\end{algorithmic}
\end{algorithm}

The total computational cost is of order $m \sum_{S' \in \mathcal{A}} 2^{|S'|}$ component evaluations per data point.
This is manageable when the selected feature sets are few and of low order, although the dependence on  the size of the feature sets remains exponential.
In practice, instead of using all data points, we use only 100 empirical quantiles of each feature to calculate the identifiable projections.
In the experiments in Section~\ref{app:cost_identification}, using 100 quantiles takes a few seconds and at most a few hundred megabytes of GPU memory for SCAN models with $1{,}000$ basis functions and up to $p=100$ inputs.
Note that the computational cost is governed by the selected feature sets rather than by the enumeration of all input feature subsets.

\begin{remark}[Example of post-hoc identification]
Let $p = 2$ and let $\nu$ be the uniform distribution on $[0,1]^2$.
Consider $f(x_1, x_2) = f_{\{1\}}(x_1) + f_{\{2\}}(x_2) + f_{\{1,2\}}(x_1, x_2)$
with $f_{\{1\}}(x_1) = x_1$, $f_{\{2\}}(x_2) = x_2$ and
$f_{\{1,2\}}(x_1, x_2) = x_1 x_2$.
Applying (\ref{eq:purify}) to the second-order component gives
\begin{align*}
\Pi_{\emptyset} f_{\{1,2\}} = \tfrac{1}{4}, 
&\Pi_{\{1\}} f_{\{1,2\}} = \tfrac{1}{2}\big(x_1 - \tfrac12\big), \\
\Pi_{\{2\}} f_{\{1,2\}} = \tfrac{1}{2}\big(x_2 - \tfrac12\big), 
&\Pi_{\{1,2\}} f_{\{1,2\}} = \big(x_1 - \tfrac12\big)\big(x_2 - \tfrac12\big).
\end{align*}
Aggregating these with
$\Pi_{\{1\}} f_{\{1\}} = x_1 - \tfrac12$,
$\Pi_{\{2\}} f_{\{2\}} = x_2 - \tfrac12$ and the constants
$\Pi_\emptyset f_{\{1\}} = \Pi_\emptyset f_{\{2\}} = \tfrac12$ yields
\begin{align*}
f^{*}_{\emptyset} = \tfrac{5}{4}, \quad
&f^{*}_{\{1\}} = \tfrac{3}{2}\big(x_1 - \tfrac12\big), \\
f^{*}_{\{2\}} = \tfrac{3}{2}\big(x_2 - \tfrac12\big), \quad
&f^{*}_{\{1,2\}} = \big(x_1 - \tfrac12\big)\big(x_2 - \tfrac12\big).
\end{align*}
Each $f^{*}_S$ with $S \neq \emptyset$ integrates to zero in each of its
arguments, and $\sum_{S} f^{*}_S = f$. 
Note that starting from the other
representation $-x_1 + x_2 + x_1(x_2 + 2)$ yields the same components.
\end{remark}

\section{Theoretical Properties of SCAN}
\label{sec:theory}

\subsection{Universal Approximation}

In this section, we establish a universal approximation theorem for SCAN, in the spirit of the classical result for standard neural networks (\cite{hornik1989multilayer}).
That is, SCAN can approximate any H\"older continuous component $f_{0,S}$ to a desired level of accuracy by increasing the number of basis functions.
This result is used in Section~\ref{sec:minimax} to derive the statistical convergence rate of SCAN for the true function $f_0$.
For simplicity, we assume $\mathcal{X} = [0,1]^p$ and consider SCAN without the global component-selection parameter $\mathbf{g}$.
For the case with $\mathbf{g}$, see Remark~\ref{remark:global_param}.

For $T \in \mathbb{N}$ and $B >0$, let $\Psi(T, B)$ denote the set of SCAN parameters $\boldsymbol{\psi} = \{ w_t, \boldsymbol{\zeta}_t : t \in [T] \}$ with $T$ basis functions such that, for all $j \in [p]$ and $t \in [T]$,
\begin{align*}
\footnotesize
\max\bigl\{ |w_t|,\ |\theta_{j,t}|,\ |\beta_{j,t,1}|,\ |\beta_{j,t,2}|,\ |b_{j,t}| \bigr\} \leq B, 
\gamma_{j,t} \in [1/B, B].
\end{align*}

\begin{theorem}[Approximation of a single component in the sup norm]\label{thm:approx}
Let $S \subseteq [p]$ with $|S| = d$ and $T = r^d$ for some $r \in \mathbb{N}$.
Assume that
\begin{enumerate}
    \item[(i)] $|f_{0,S}(\mathbf{x}_S) - f_{0,S}(\mathbf{y}_S)| \leq L \Vert \mathbf{x}_S - \mathbf{y}_S \Vert_{2}^{\rho}$ for all $\mathbf{x}_S, \mathbf{y}_S \in [0,1]^d$, for some $\rho \in (0,1]$ and $L > 0$,
    \item[(ii)] $\Vert f_{0,S} \Vert_{\infty} \leq F$ for some $F > 0$.
\end{enumerate}
Fix the truncation threshold $0<\epsilon_{\alpha}<1/d.$
Then, there exists
$\boldsymbol{\psi} \in \Psi(T, B)$  with
\begin{align*}
B = \max\Bigl\{ &2,\; \tfrac{1}{\mathbb{T}},\; 2^d F,\; 2pT^{1/d}, 40d\,T^{2/d},\\
&\tfrac{\mathbb{T}}{2}\log\max\bigl\{\tfrac{1}{d\epsilon_\alpha}, \tfrac{(p-d)\epsilon_\alpha}{1-d\epsilon_\alpha}\bigr\} + 1 \Bigr\}
\end{align*}
such that
\begin{small}
\begin{align}
\sup_{\mathbf{x} \in \mathcal{X}} \big| f_{0,S}(\mathbf{x}_S) - f_{\text{SCAN}}(\mathbf{x} \mid \boldsymbol{\psi},\epsilon_{\alpha}) \big|
\leq \frac{4^{d} L d^{\rho/2}}{T^{\rho/d}} + 2^{d+1} F\, T e^{-T^{1/d}/2},
\label{eq:approx_bound}
\end{align}
\end{small}
and the feature sets $S_t$s of each basis $\mathcal{M}_t$ in $f_{\text{SCAN}}(\mathbf{x} \mid \boldsymbol{\psi},\epsilon_{\alpha})$ are all equal to $S.$
\end{theorem}

Theorem~\ref{thm:approx} shows that SCAN can approximate any H\"older continuous component to an arbitrary precision by choosing $T$ sufficiently large.
Since the bound in (\ref{eq:approx_bound}) is of order $T^{-\rho/d}$, the approximation error decays at the rate $T^{-\rho/d}$, and thus the number of basis functions required for a given precision increases as $d$ increases.
In addition, if $f_0$ has a sparse functional ANOVA decomposition with the maximum order $d$
and all signal components are  $\rho$-H\"older continuous, we can show
the same upper bound for $f_0$ by approximating each component separately.

\subsection{Minimax Optimality}
\label{sec:minimax}

We consider the regression model $Y = f_0(\mathbf{X}) + \varepsilon$ with $\mathbb{E}(\varepsilon \mid \mathbf{X}) = 0$.
For a given truncation threshold $\epsilon_{\alpha} > 0$, define the class of SCAN functions with bounded parameters as
\begin{align*}
\mathcal{F}(T, B, \epsilon_{\alpha}) := \big\{ f_{\text{SCAN}}(\cdot \mid \boldsymbol{\psi},\epsilon_{\alpha}) : \boldsymbol{\psi} \in \Psi(T, B) \big\}.
\end{align*}
For given i.i.d. observations $\{(\mathbf{X}_i,Y_i)\}_{i=1}^n$, let
\begin{align*}
\hat{f}^{\mathrm{raw}}_n \in \operatorname*{argmin}_{f \in \mathcal{F}(T_n, B_n, \epsilon_{\alpha})}
\frac{1}{n} \sum_{i=1}^{n} \big( Y_i - f(\mathbf{X}_i) \big)^{2},
\:\:
\hat f_n := \mathcal{T}_{\beta_n} \hat f^{\mathrm{raw}}_n,
\end{align*}
where $\mathcal{T}_{\beta}(z) := \max\{-\beta, \min\{\beta, z\}\}$, $T_n$, $B_n$, and $\epsilon_{\alpha}$ denote the number of basis functions, the parameter bound, and the truncation threshold, respectively, and $\beta_n = c_1 \log n$ for some constant $c_1 > 0$.

\begin{assumption}\label{assump:minimax}
$f_0(\mathbf{x}) = \sum_{S \in \mathcal{S}} f_{0,S}(\mathbf{x}_S)$ with $\mathcal{S} \subseteq \mathcal{P}([p])$ and $\max_{S \in \mathcal{S}} |S| = d$ , where each $f_{0,S}$ satisfies conditions (i)-(ii) of Theorem~\ref{thm:approx}. 
Moreover, $\mathbb{E}\{\exp(c_0 Y^2)\} < \infty$ for some $c_0 > 0$.
\end{assumption}

\begin{theorem}[Minimax-optimal convergence rate]\label{thm:conv_rate}
Under Assumption~\ref{assump:minimax}, let
\begin{align*}
T_n = \lceil c_T\, n^{d/(2\rho+d)} \rceil
\end{align*}
for a constant $c_T \geq 2^{d} |\mathcal{S}|$,
$B_n = n^{\kappa}$ for some $\kappa > 2/(2\rho+d)$, $\beta_n = c_1 \log n$ for some $c_1 > 0$, and fix $\epsilon_\alpha \in (0, 1/d)$. Then
\begin{align*}
\mathbb{E} \Vert \hat{f}_{n} - f_0 \Vert_{2,\mu}^{2} \leq C (\log n)^{3}\, n^{-\frac{2\rho}{2\rho + d}}
\end{align*}
for some constant $C > 0$, which is independent of $n$.
\end{theorem}

Theorem~\ref{thm:conv_rate} gives a minimax optimal convergence rate of $n^{-2\rho/(2\rho+d)}$ up to a logarithmic factor for the least squares estimator under Assumption~\ref{assump:minimax}.
The order of upper bound depends on the maximum order $d$ of the signal interactions rather than the input feature dimension $p$.
Thus, for fixed dimension and a sparse collection of low-order components, the estimator attains the rate associated with the maximum order of signal interactions.

\begin{remark}[Global component-selection parameter]
\label{remark:global_param}
For simplicity, the results in this section are stated without the global component-selection parameter $\mathbf{g}$ in (\ref{eq:dual_temp}), but they remain valid with $\mathbf{g}$, provided that $\mathbf{g}$ is bounded in the same way as the other parameters, i.e., $\Vert \mathbf{g} \Vert_{\infty} \leq B$.
Indeed, writing $\boldsymbol{\Theta}''_t := \boldsymbol{\Theta}_t + (\mathbb{T}/\mathbb{T}_g)\, \mathbf{g}$, we have $\text{softmax}(\boldsymbol{\Theta}_t/\mathbb{T} + \mathbf{g}/\mathbb{T}_g) = \text{softmax}(\boldsymbol{\Theta}''_t/\mathbb{T})$ with $\Vert \boldsymbol{\Theta}''_t \Vert_{\infty} \leq B(1 + \mathbb{T}/\mathbb{T}_g)$.
Hence, the SCAN functions with $\mathbf{g}$ form a subset of the SCAN functions without $\mathbf{g}$ whose parameter bound is enlarged from $B$ to $B(1 + \mathbb{T}/\mathbb{T}_g)$.
Since the temperatures are fixed, the factor $1 + \mathbb{T}/\mathbb{T}_g$ is a constant independent of $n$, and thus, by Lemma~\ref{lem:covering}, the complexity of the class is of the same order.
Conversely, taking $\mathbf{g} = \mathbf{0}$ recovers the SCAN functions without $\mathbf{g}$,
so that the approximation results also hold.
Thus, the same rate applies when the global component-selection parameter is included and the temperatures are fixed.
\end{remark}

\begin{table*}[h]
\centering
\providecommand{\res}[2]{#1~(#2)}
\caption{\footnotesize \textbf{Prediction performance on nine benchmark datasets.} Entries are means (standard deviations) over ten trials.
$\downarrow (\uparrow)$ means that lower (higher) is better.
The best results among the functional ANOVA models are in \textbf{bold}.}
\label{table:pred_perf}
\setlength{\tabcolsep}{4pt}
\resizebox{\textwidth}{!}{%
\begin{tabular}{l|cccc|ccccc}
\toprule
 & \multicolumn{4}{c|}{Regression (RMSE $\downarrow$)} & \multicolumn{5}{c}{Classification (AUROC $\uparrow$)} \\
\cmidrule(lr){2-5} \cmidrule(lr){6-10}
Model & \textsc{Calhousing} & \textsc{Wine} & \textsc{Microsoft} & \textsc{Yahoo} & \textsc{Churn} & \textsc{Letter} & \textsc{Credit} & \textsc{Drybean} & \textsc{Madelon} \\
\midrule \midrule
\multicolumn{10}{l}{\textit{Functional ANOVA models without component selection}} \\
\midrule
NAM (\cite{agarwal2021neural}) & \res{0.525}{0.02} & \res{0.720}{0.02} & \res{0.774}{0.001} & \res{0.797}{0.002} & \res{0.846}{0.01} & \res{0.986}{0.001} & \res{0.980}{0.01} & \res{0.996}{0.001} & \res{0.644}{0.011} \\
NBM (\cite{radenovic2022neural})& \res{0.502}{0.03} & \res{0.702}{0.03} & \res{0.770}{0.001} & \res{0.783}{0.002} & \res{0.845}{0.01} & \res{0.990}{0.001} & \res{\textbf{0.985}}{0.01} & \res{0.995}{0.001} & \res{0.670}{0.031} \\
ANOVA-TPNN (\cite{park2025tensor})& \res{0.512}{0.01} & \res{0.704}{0.02} & \res{0.756}{0.001} & \res{0.787}{0.002} & \res{0.842}{0.012} & \res{0.984}{0.001} & \res{0.984}{0.001} & \res{\textbf{0.998}}{0.001} & \res{0.587}{0.013} \\
\midrule
\multicolumn{10}{l}{\textit{Functional ANOVA models with component selection}} \\
\midrule
NODE-GAM (\cite{chang2021node})& \res{0.515}{0.01} & \res{0.723}{0.02} & \res{0.756}{0.001} & \res{0.777}{0.001} & \res{0.830}{0.011} & \res{0.988}{0.001} & \res{\textbf{0.985}}{0.006} & \res{0.996}{0.001} & \res{0.792}{0.029} \\
EBM (\cite{nori2019interpretml})& \res{0.473}{0.01} & \res{\textbf{0.663}}{0.01} & \res{0.759}{0.001} & \res{0.775}{0.001} & \res{0.846}{0.013} & \res{0.988}{0.001} & \res{0.979}{0.008} & \res{0.996}{0.001} & \res{0.818}{0.022} \\
SCAN (Ours) & \res{\textbf{0.471}}{0.01} & \res{0.690}{0.02} & \res{\textbf{0.744}}{0.001} & \res{\textbf{0.763}}{0.002} & \res{\textbf{0.848}}{0.01} & \res{\textbf{0.995}}{0.001} & \res{0.983}{0.01} & \res{0.997}{0.001} & \res{\textbf{0.895}}{0.018} \\
\midrule \midrule
\multicolumn{10}{l}{\textit{Black-box models}} \\
\midrule
XGB (\cite{chen2016xgboost})& \res{0.452}{0.01} & \res{0.635}{0.03} & \res{0.712}{0.001} & \res{0.731}{0.001} & \res{0.846}{0.01} & \res{0.997}{0.001} & \res{0.983}{0.001} & \res{0.997}{0.001} & \res{0.894}{0.011} \\
MLP (\cite{hornik1989multilayer})& \res{0.518}{0.01} & \res{0.696}{0.01} & \res{0.741}{0.002} & \res{0.768}{0.003} & \res{0.842}{0.01} & \res{0.980}{0.001} & \res{0.980}{0.001} & \res{0.997}{0.001} & \res{0.605}{0.024} \\
KAN (\cite{liu2024kan})& \res{0.505}{0.01} & \res{0.725}{0.01} & \res{0.750}{0.001} & \res{0.765}{0.001} & \res{0.847}{0.01} & \res{0.996}{0.001} & \res{0.982}{0.001} & \res{0.997}{0.001} & \res{0.616}{0.008} \\
\bottomrule
\end{tabular}%
}
\end{table*}

\subsection{Component Selection Consistency}
\label{sec:convergence}

Let $\bar{\Pi}_S$ denote the population counterpart of $\Pi_S$, obtained by replacing $\nu$ with $\bar{\mu}$ in (\ref{eq:purify}).
We define the \emph{identified true components} and the \emph{identified component index set} as
\begin{align*}
f^{*}_{0,S} := \bar{\Pi}_S f_0,
\qquad
\mathcal{S}^{*} := \{ S \neq \emptyset : f^{*}_{0,S} \neq 0 \},
\end{align*}
and write $r_n := (\log n)^{3}\, n^{-\frac{2\rho}{2\rho + d}}$.

\begin{theorem}[Component selection consistency]\label{cor:selection}
Suppose that the conditions of Theorem~\ref{thm:conv_rate} hold and that $\mathbf{X}$ has a density $p_X$ on $\mathcal{X}$ satisfying $p_L \leq p_X(\mathbf{x}) \leq p_U$ for all $\mathbf{x} \in \mathcal{X}$ and for some constants $0 < p_L \leq p_U < \infty$.
For a threshold $\tau_{f,n} > 0$, let
\begin{align*}
\hat{\mathcal{S}}_n := \big\{ S \neq \emptyset : \Vert \Pi_S \hat f_n \Vert_{2,n} > \tau_{f,n} \big\}
\end{align*}
be the set of selected components.
If $\tau_{f,n} \to 0$ and $\tau_{f,n}^{2} / r_n \to \infty$, then
\begin{align*}
\mathbb{P}\big( \hat{\mathcal{S}}_n = \mathcal{S}^{*} \big) \to 1 \quad \text{as} \quad n \to \infty.
\end{align*}
\end{theorem}

Theorem~\ref{cor:selection} shows that the true components can be consistently selected by thresholding the empirical $L_2$ norms of the identified components by SCAN.
Motivated by this result, we use $\Vert \Pi_S \hat f_n \Vert_{2,n}$ as the importance score of the component $S$.
Related experiments are reported in Section~\ref{sec:comp_selc}.

Moreover, under a mild regularity condition, the identified true components are contained in the downward closure of the discovered components with probability tending to one, i.e., $\mathbb{P}\big( \mathcal{S}^{*} \subseteq \mathcal{D}(\hat{\mathcal{A}}_n) \big) \to 1$ (see Proposition~\ref{prop:screening} in Section~\ref{app:screening}), where $\hat{\mathcal{A}}_{n}$ is the set of feature sets discovered by SCAN.
That is, restricting the post-hoc identification to $\mathcal{D}(\hat{\mathcal{A}}_n)$ loses none of the identified true components with probability tending to one.

We note that the argument of Theorem~\ref{cor:selection} relies only on the convergence rate in Theorem~\ref{thm:conv_rate}, and hence the same selection consistency holds for any estimator that attains such a rate.
Nevertheless, SCAN has two advantages.
First, its convergence rate depends only on the maximum order $d$ of the components and not on the input feature dimension $p$,
so that the threshold $\tau_{f,n}$ can decrease faster and the true components can be recovered with fewer samples than with estimators whose rates deteriorate with $p$.
Second, SCAN explicitly discovers a set of low-order candidate components $\hat{\mathcal{A}}_n$ during training,
from which the identified true components can be recovered.
This is practically important, since post-hoc identification is computationally demanding when the components involve high-order interactions (see Appendix M of \cite{park2025tensor}).
Hence, it is desirable that the maximum order of the components in $\hat{\mathcal{A}}_n$,
$\max_{S \in \hat{\mathcal{A}}_n} |S|$, be as small as possible,
while $\mathcal{D}(\hat{\mathcal{A}}_n)$ still contains $\mathcal{S}^{*}$.
Empirical results showing that the components in $\hat{\mathcal{A}}_n$ discovered by SCAN have low maximum order
are given in Section~\ref{app:structural_order}.

\begin{remark}[Choice of the truncation threshold]\label{rem:eps_choice}
The results in this section hold for any fixed truncation threshold $\epsilon_\alpha \in (0, 1/d)$.
The threshold affects the constants through the parameter bound and need not decrease with $n$.
Since the maximum order $d$ is unknown in practice, a small fixed threshold allows a broader range of component orders.
Unless stated otherwise, we use $\epsilon_\alpha=0.01$ for truncation.
This setting satisfies the theoretical threshold condition when the true components have orders less than $100$.
\end{remark}

\section{Experiments}
\subsection{Experimental Settings}
$\newline$
\textbf{Baseline models.}
We compare SCAN with NAM (\cite{agarwal2021neural}), NBM (\cite{radenovic2022neural}), ANOVA-TPNN (\cite{park2025tensor}), EBM (\cite{nori2019interpretml}), and NODE-GAM (\cite{chang2021node}) as well as
the black-box references including XGB (\cite{chen2016xgboost}), MultiLayer Perceptrons (MLP, \cite{hornik1989multilayer}), and KAN (\cite{liu2024kan}).
Hyperparameter settings are given in Section \ref{app:hyper}. \\
$\newline$
\textbf{Implementation details.}
For XGB, EBM, MLP, and KAN, we standardize the continuous input features and the target variable for regression.
For NAM, NBM and SCAN, the min-max scaling is applied to the input features.
For NODE-GAM and ANOVA-TPNN, the quantile transformation is applied to the input features.
For NAM, NBM, NODE-GAM, and ANOVA-TPNN, the maximum component order is selected from $\{1,2\}$.
\\
$\newline$
\textbf{Datasets.}
We evaluate prediction performance on nine benchmark datasets, including \textsc{Calhousing} (\cite{calhousing}), \textsc{Wine} (\cite{wine_quality}), \textsc{Microsoft} (\cite{micro}), \textsc{Yahoo} (\cite{yahoo}), \textsc{Churn} (\cite{churn}), \textsc{Letter} (\cite{Letter}), \textsc{Credit} (\cite{credit}), \textsc{Drybean} (\cite{drybean}), \textsc{Madelon} (\cite{madelon_171}), and  Breast cancer gene (\cite{hatzis2011genomic}).
More details are in Section \ref{app:data}.

\begin{table}[h]
\centering
\caption{\footnotesize \textbf{Training time per epoch (seconds) and peak GPU memory (GB) as the number of input features $p$ increases.} OOM indicates that the model ran out of GPU memory (24GB).}
\label{tab:computation}
\begin{tabular}{l|ccccc}
\toprule
 & \multicolumn{5}{c}{$p$} \\
\cmidrule(lr){2-6}
Model & 10 & 50 & 100 & 200 & 300 \\
\midrule
\multicolumn{6}{l}{\textit{Training time per epoch (sec)}} \\
\midrule
SCAN            & \textbf{0.25} & \textbf{0.25} & \textbf{0.29} & \textbf{0.41}   & \textbf{0.45} \\
NODE-GA$^{2}$M         & 1.66 & 1.72 & 1.75 & 1.85   & 1.88 \\
NB$^{2}$M       & 0.39          & 2.6           & 10.0          & OOM             & OOM \\
NA$^{2}$M       & 0.59          & 4.7           & 20.9          & 98.0            & OOM \\
ANOVA-T$^{2}$PNN & 3.9          & 98.1          & 397.7         & 1{,}801         & 3{,}718 \\

\midrule
\multicolumn{6}{l}{\textit{Peak GPU memory (GB)}} \\
\midrule
SCAN            & \textbf{0.03} & \textbf{0.07} & \textbf{0.09} & \textbf{0.11}   & \textbf{0.11} \\
NODE-GA$^{2}$M         & 0.29 & 0.31 & 0.35 & 0.38   & 0.40 \\
NB$^{2}$M       & 0.13          & 2.65          & 10.48         & OOM             & OOM \\
NA$^{2}$M       & 0.05          & 0.72          & 2.82          & 11.17           & OOM \\
ANOVA-T$^{2}$PNN & \textbf{0.03} & 0.28         & 1.05          & 4.16            & 9.34 \\
\bottomrule
\end{tabular}
\end{table}

\subsection{Prediction Performance}
\label{sec:pred}
In this section, we compare the prediction performance of SCAN with the baseline models, including NAM, NBM, ANOVA-TPNN, EBM, NODE-GAM, XGB, MLP, and KAN.
Each dataset is randomly divided into train, validation, and test datasets in a 70/10/20 ratio.
The validation dataset is used to select the optimal epoch and hyperparameters, and the test dataset is used to evaluate the prediction performance of the trained models.
We repeat this random split 10 times and report the averages and standard deviations of the prediction performance.
As a performance measure, we use the Root Mean Square Error (RMSE) for regression problems and the Area Under the ROC Curve (AUROC) for classification problems.

Table~\ref{table:pred_perf} reports the averages (standard deviations) of the prediction performance over 10 trials.
The results of NAM, NBM, and ANOVA-TPNN are taken from \cite{park2025tensor}.
For NODE-GAM, the results on \textsc{Microsoft} and \textsc{Yahoo} are obtained from our own experiments, while those on the other datasets are taken from \cite{park2025tensor}.
All the remaining results are obtained from our own experiments.
SCAN achieves competitive results on all nine datasets.
On regression tasks, SCAN achieves the best performance among the functional ANOVA models on three of the four datasets, and on classification tasks, it performs better than or comparably to all baselines, including XGB.
In particular, SCAN outperforms the functional ANOVA models restricted to main effects and pairwise interactions on most datasets, which suggests the benefit of incorporating higher-order components.

\subsection{Global Component-selection Parameter Annealing}
\label{sec:exp_madelon}

To examine the effect of the global component-selection parameter $\mathbf{g}$ and the temperature annealing introduced in Section~\ref{sec:global_prior}, we conduct experiments on \textsc{Madelon} dataset.
\textsc{Madelon} is a synthetic binary classification dataset with 500 features, among which only a few feature are relevant and the remaining features are noisy features.
Moreover, it is known that there exist many higher-order signal interactions (\cite{park2025bayesian}).
Hence, \textsc{Madelon} is a challenging benchmark for the functional ANOVA models, since a model must identify a small number of relevant features among many noisy ones and estimate higher-order interactions among them.

We compare the following three variants of SCAN:
(i) SCAN with a fixed temperature $\mathbb{T}$ and without $\mathbf{g}$,
(ii) SCAN with the annealing of $\mathbb{T}$ only and without $\mathbf{g}$, and
(iii) SCAN with $\mathbf{g}$ and the annealing of both $\mathbb{T}$ and $\mathbb{T}_g$.
For all variants, we use $T = 200$ basis functions, a learning rate of $10^{-2}$, and a weight decay of $10^{-3}$.
The final temperatures are selected by validation AUROC.

Table~\ref{table:madelon} shows that SCAN with a fixed local temperature achieves an AUROC of 0.656.
Annealing the local temperature increases the AUROC to 0.785.
Adding the global component-selection parameter together with global temperature annealing gives an AUROC of 0.895, close to XGB's 0.894.
This comparison supports the necessity of the global component-selection parameter and annealing of the temperature parameters.

\begin{table}[h]
\centering
\caption{\footnotesize \textbf{Effect of the global component-selection parameter and
temperature annealing in SCAN on \textsc{Madelon} dataset.} We report the averages (standard deviations) of the AUROC.}
\label{table:madelon}
\begin{tabular}{lccc}
\toprule
 & Fixed $\mathbb{T}$ & Annealing $\mathbb{T}$ & Annealing $\mathbb{T}$ and $\mathbb{T}_g$ \\
\midrule
AUROC $\uparrow$ & 0.656 (0.037) & 0.785 (0.036) & \textbf{0.895} (0.018) \\
\bottomrule
\end{tabular}
\end{table}

\subsection{Component Selection}
\label{sec:comp_selc}
\begin{table*}[t]
\centering
\providecommand{\res}[2]{#1~(#2)}
\caption{\textbf{Performance of component selection on synthetic datasets.} Entries are means (standard deviations) of the AUROC over five repetitions.}
\label{tab:component_selection}
\setlength{\tabcolsep}{4pt}
\resizebox{\textwidth}{!}{%
\begin{tabular}{l ccc ccc ccc}
\toprule
True model & \multicolumn{3}{c}{$f^{(1)}$} & \multicolumn{3}{c}{$f^{(2)}$} & \multicolumn{3}{c}{$f^{(3)}$} \\
\cmidrule(lr){2-4} \cmidrule(lr){5-7} \cmidrule(lr){8-10}
Order & 1 & 2 & 3 & 1 & 2 & 3 & 1 & 2 & 3 \\
\midrule
NAM        & \res{0.528}{0.023} & \res{0.508}{0.024} & --- & \res{0.417}{0.015} & \res{0.838}{0.009} & --- & \res{0.522}{0.011} & \res{0.542}{0.017} & --- \\
NBM        & \res{0.737}{0.059} & \res{0.739}{0.053} & --- & \res{0.660}{0.092} & \res{0.950}{0.027} & --- & \res{0.708}{0.071} & \res{0.726}{0.039} & --- \\
ANOVA-TPNN & \res{0.949}{0.049} & \res{0.893}{0.042} & --- & \res{0.808}{0.094} & \res{0.946}{0.004} & --- & \res{0.763}{0.069} & \res{0.558}{0.044} & --- \\
NODE-GAM   & \res{0.954}{0.026} & \res{0.805}{0.056} & --- & \res{0.857}{0.036} & \res{0.950}{0.000} & --- & \res{0.851}{0.028} & \res{0.702}{0.036} & --- \\
EBM        & \res{0.930}{0.029} & \res{0.805}{0.049} & --- & \res{0.848}{0.066} & \res{0.950}{0.000} & --- & \res{0.862}{0.052} & \res{0.796}{0.021} & --- \\
\midrule
SCAN (Ours) & \res{\textbf{0.983}}{0.022} & \res{\textbf{0.927}}{0.019} & \res{\textbf{0.937}}{0.016} & \res{\textbf{0.956}}{0.030} & \res{\textbf{0.999}}{0.001} & \res{\textbf{1.000}}{0.000} & \res{\textbf{0.971}}{0.019} & \res{\textbf{0.860}}{0.041} & \res{\textbf{0.810}}{0.086} \\
\bottomrule
\end{tabular}%
}
\end{table*}

We evaluate whether SCAN correctly identifies the true components by analyzing synthetic datasets.
We generate data from three simulation models $f^{(1)}$, $f^{(2)}$, and $f^{(3)}$, whose forms and input feature distributions are explained in Section \ref{app:syn_data}.

Let $\Pi_S \hat f$ denote the identified component of SCAN for $S \subseteq [p]$,
obtained by applying the post-hoc identification in Section~\ref{sec:identification}.
To quantify how much each component contributes to the prediction, we define the importance of $\Pi_S \hat f$ as its empirical $L_2$ norm,
\begin{align}
\mathrm{Imp}(S) := \Vert \Pi_S \hat f \Vert_{2,n} = \bigg( \frac{1}{n} \sum_{i=1}^{n} \big(\Pi_S \hat f(\mathbf{x}_{i,S})\big)^{2} \bigg)^{1/2},
\label{eq:importance}
\end{align}
where $\mathbf{x}_1, \ldots, \mathbf{x}_n$ are given data.
A larger $\mathrm{Imp}(S)$ indicates that the component $\Pi_{S}\hat{f}$ contributes more to the prediction.
For each order $k \in \{1, 2, 3\}$, we evaluate how well the importance scores identify the true components by the AUROC, where the signal components are labeled
 as positive and the non-signal components as negative.
A higher AUROC indicates that the signal components have larger important scores than the non-signal ones.

We repeat the experiments 5 times and report the averages and standard deviations.
Table~\ref{tab:component_selection} reports the component selection AUROC up to the third order interactions.
Since ANOVA-TPNN, NAM, NBM, EBM and NODE-GAM are restricted to components up to the second order due to their computational cost, we report their component selection performance only up to the second order.
SCAN gives the highest component selection AUROC in all settings.
Moreover, SCAN recovers higher-order components well in these settings.
Further experimental details are given in Section~\ref{app:syn_data}.

\subsection{Computational Efficiency}
\label{sec:exp_computation}

To evaluate the scalability of SCAN with respect to the input feature dimension, we compare the training cost of SCAN with the existing neural functional ANOVA models that estimate components up to the second order, namely NB$^{2}$M, NA$^{2}$M, ANOVA-T$^{2}$PNN and NODE-GA$^{2}$M.
We generate synthetic datasets with $n = 10{,}000$ samples and vary the number of input features $p \in \{10, 50, 100, 200, 300\}$.
All models are trained with a batch size of 256, giving 40 steps per epoch, on a single GPU with 24 GB of memory.
We measure the training time per epoch and the peak GPU memory.
We set the number of basis functions in SCAN to $T = 200$.
For NODE-GA$^{2}$M, we set 200 trees, 4 layers and tree depth 5.
NB$^{2}$M uses 200 shared basis functions computed by an MLP with hidden layers of sizes (256, 128, 128), and NA$^{2}$M uses a separate MLP with hidden layers of sizes (64, 64, 32) for each component.
For ANOVA-T$^{2}$PNN, we use $K = 10$ basis functions for each component.

Table~\ref{tab:computation} summarizes the results.
SCAN has the shortest training time and the smallest (or tied smallest) peak GPU memory for all cases.
As $p$ increases from 10 to 300, the training time of SCAN increases only from 0.25 to 0.45 seconds per epoch, and its peak memory from 0.03 to 0.11 GB.
In contrast, the costs of NB$^{2}$M, NA$^{2}$M, and ANOVA-T$^{2}$PNN grow rapidly with $p$, since the number of second-order components increases quadratically in $p$.
NB$^{2}$M requires 10.48 GB of memory at $p = 100$ and runs out of memory for $p \geq 200$.
NA$^{2}$M runs up to $p = 200$, but requires 98 seconds per epoch, and runs out of memory at $p = 300$.
ANOVA-T$^{2}$PNN does not run out of memory, but its training time increases sharply, reaching approximately 30 minutes per epoch at $p = 200$ and one hour at $p = 300$.
Even though NODE-GA$^{2}$M only estimates components of order at most two, it is four to seven times slower and uses three to ten times more memory than SCAN for every $p$.
In contrast, SCAN can estimate components of any order.

\begingroup
\emergencystretch=1em
\interfootnotelinepenalty=10000
\renewcommand{\thefootnote}{\textcolor{teal}{\arabic{footnote}}}
\section{Application to High-Dimensional Gene-Expression Dataset}
\label{sec:applications}

In this section, we analyze high-dimensional gene expression data using SCAN to illustrate that SCAN can facilitate new scientific discoveries.
\medskip

\textbf{Breast cancer gene-expression dataset.}
Gene-expression data typically contain thousands of genes measured on a small number of patients,
and it is of interest to identify which genes, or interactions among them, explain the prediction.
We apply SCAN to the breast cancer gene-expression data of~\cite{hatzis2011genomic}.
The data consist of pretreatment microarray measurements, which we convert to gene-level values by taking the median of the probes mapped to each gene.
The response indicates whether a patient achieves a pathological complete response (pCR), that is, no residual invasive cancer after neoadjuvant chemotherapy.
Following the original study, we use the development cohort (306 patients, 57 with pCR) for training and the external validation cohort (182 patients, 42 with pCR) only for evaluation.
All $p = 12{,}502$ genes are used as inputs without preliminary screening, which yields about 78 million possible second-order components.
\\
$\newline$
\textbf{Experimental setup.}
We compare SCAN with NODE-GAM and EBM.
Each model is trained on the development cohort with five different random seeds and evaluated on the external cohort, and the results are reported in Table~\ref{tab:application_breast_prediction}.
Further details are given in Section S.B.4 of Supplementary Material.
NAM, NBM, and ANOVA-TPNN could not be run under our computational resources, since they would need to estimate an excessively large number of components.

\begin{table}[htbp]
\centering
\caption{\footnotesize \textbf{Prediction performance on the external breast cancer cohort.}
AUROC is larger for better prediction. The best reported values are in bold. A hyphen denotes an experiment not run at this input dimension.}
\label{tab:application_breast_prediction}
\resizebox{\linewidth}{!}{%
\begin{tabular}{lccccccc}
\toprule
 & NAM & NBM & ANOVA-TPNN & EBM & NODE-GAM  & SCAN (ours)  \\
\midrule
AUROC $\uparrow$    & --- & --- & --- & 0.714 (0.02) & 0.654 (0.04) & \textbf{0.726} (0.02)  \\
\bottomrule
\end{tabular}%
}
\end{table}

\begin{figure*}[t]
\centering
\includegraphics[width=0.9\textwidth]{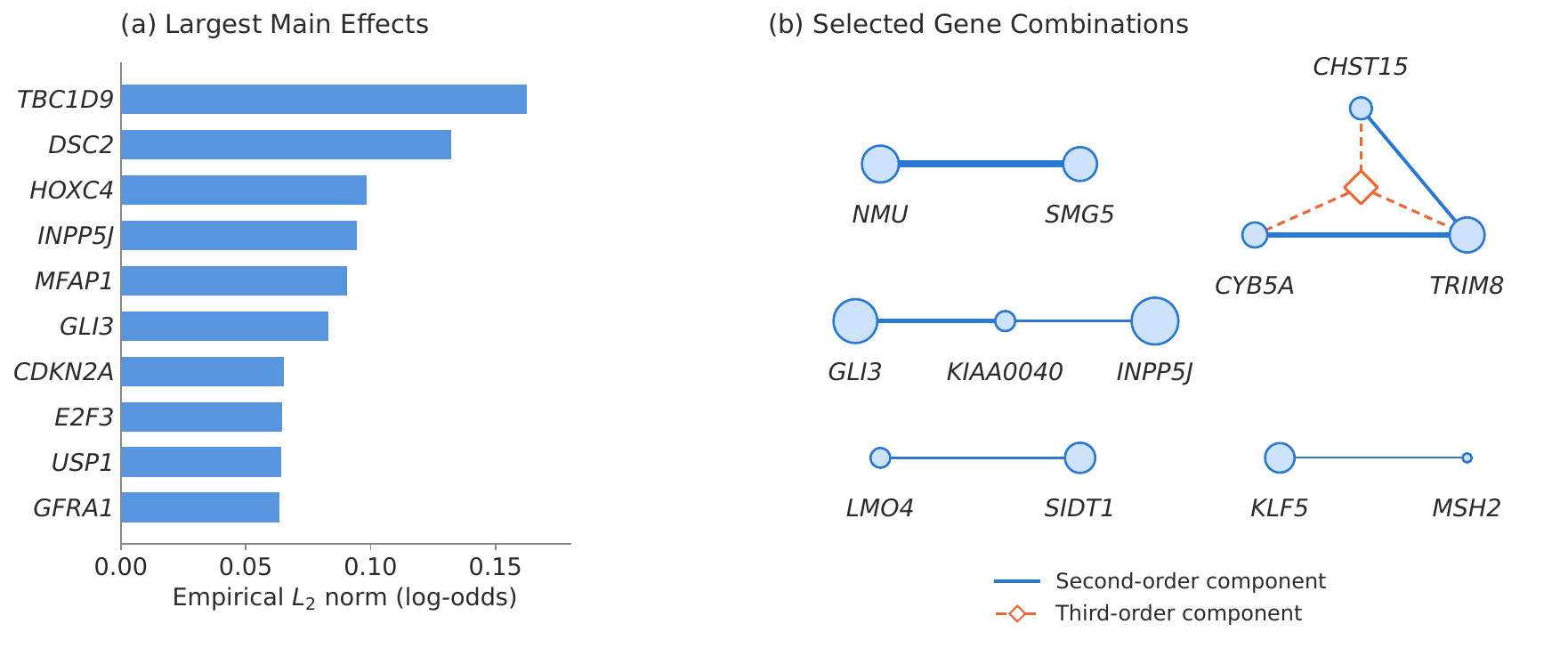}
\par\smallskip
{\small (c) Estimated Component Functions\par}
\includegraphics[width=0.9\textwidth]{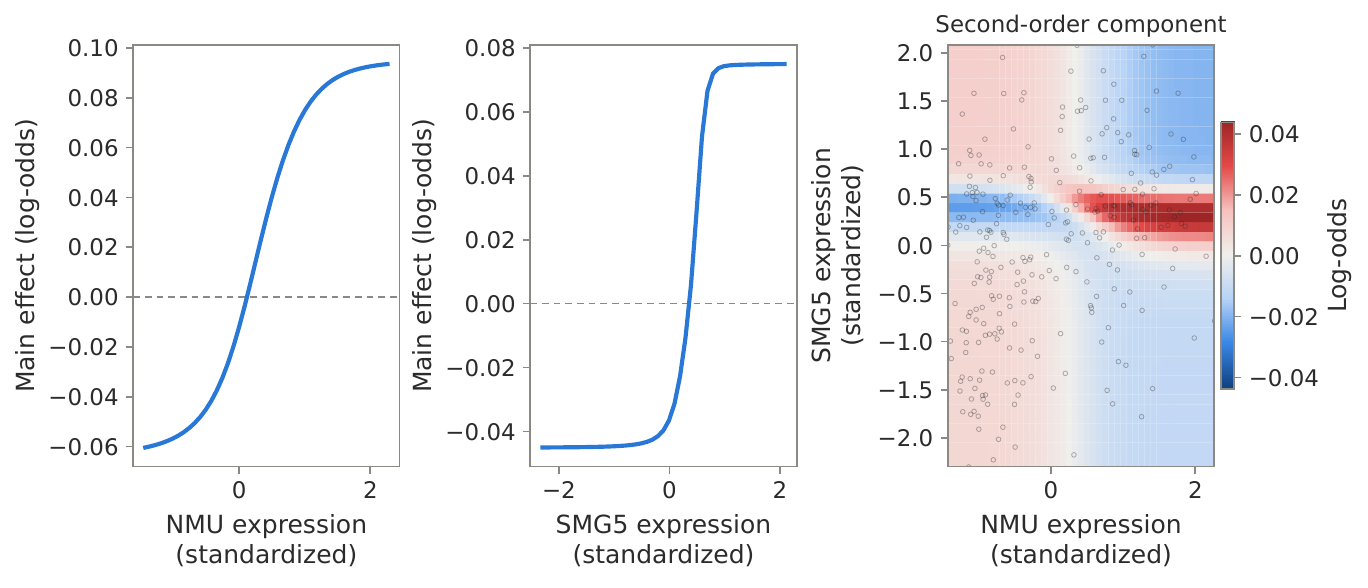}
\caption{\footnotesize \textbf{Identified SCAN components for pCR prediction.}
(a) Ten largest main effects by empirical $L_2$ norm.
(b) Retained second-order components (solid edges) and the third-order component (diamond, dashed edges); circle area and edge width are proportional to importance. Only the 12 genes involved in interactions are shown.
(c) Main effects of \textit{NMU} and \textit{SMG5} and their second-order component on the log-odds scale.}
\label{fig:application_breast}
\end{figure*}

$\newline$
\textbf{Prediction performance.}
Table~\ref{tab:application_breast_prediction} reports prediction performance on the external cohort.
The prediction performance of SCAN outperforms the baseline models.
\\
$\newline$
\textbf{Interpretation.}
We next examine the identified components of SCAN, including main effects of individual genes and interactions among genes, that contribute to its predictions.
The selected representation contains 61 components involving 53 genes, less than 0.5\% of the input variables.
These comprise 53 main effects, seven second-order components, and one third-order component.
Figure~\ref{fig:application_breast} summarizes the important main effects and all retained gene combinations.
The main effects of \textit{TBC1D9}, \textit{DSC2}, and \textit{HOXC4} have the largest empirical $L_2$ norms.
The selected interactions connect 12 genes in five groups.
The strongest second-order component involves \textit{NMU} and \textit{SMG5}.
Other groups include \textit{GLI3} and \textit{INPP5J}, each linked to \textit{KIAA0040}, and \textit{CHST15}, \textit{CYB5A}, and \textit{TRIM8}, which enter a third-order component as well as two selected pairs.
This representation reveals which combinations of genes jointly affect the response.
Figure~\ref{fig:application_breast}(c) shows the estimated components of \textit{NMU} and \textit{SMG5}.
The estimated main effects of both genes increase with their expression levels.
The second-order component is positive when \textit{NMU} expression is high and the standardized \textit{SMG5} expression is near 0.4, indicating a localized joint contribution beyond the main effects.
In this way, examining the estimated components reveals not only which genes and gene combinations contribute to the response, but also the direction and shape of their contributions.



\FloatBarrier
\endgroup

\section{Conclusion}

We proposed SCAN, a neural functional ANOVA model that learns the feature set of each basis function and thereby selects and estimates components of arbitrary order without enumerating candidate interactions.
SCAN has the universal approximation property, attains the minimax optimal rate up to a logarithmic factor, and selects the true components consistently.
Empirically, SCAN predicts better than existing neural functional ANOVA models on most benchmark datasets, recovers higher-order components, and trains at a much lower cost than models that enumerate second-order components.
It also applies directly to gene-expression data with 12,502 variables, where models that enumerate components could not be run.
However, SCAN does not satisfy the identifiability condition during training and relies on post-hoc identification to obtain uniquely defined components.
Therefore, developing a model that satisfies the identifiability condition during training is a promising direction for future work.


\bibliographystyle{IEEEtran}
\bibliography{references}

@article{agarwal2021neural,
  title={Neural additive models: Interpretable machine learning with neural nets},
  author={Agarwal, Rishabh and Melnick, Levi and Frosst, Nicholas and Zhang, Xuezhou and Lengerich, Ben and Caruana, Rich and Hinton, Geoffrey E},
  journal={Advances in neural information processing systems},
  volume={34},
  pages={4699-4711},
  year={2021}
}

@article{radenovic2022neural,
  title={Neural basis models for interpretability},
  author={Radenovic, Filip and Dubey, Abhimanyu and Mahajan, Dhruv},
  journal={Advances in Neural Information Processing Systems},
  volume={35},
  pages={8414-8426},
  year={2022}
}

@article{chang2021node,
  title={Node-gam: Neural generalized additive model for interpretable deep learning},
  author={Chang, Chun-Hao and Caruana, Rich and Goldenberg, Anna},
  journal={arXiv preprint arXiv:2106.01613},
  year={2021}
}

@article{popov2019neural,
  title={Neural oblivious decision ensembles for deep learning on tabular data},
  author={Popov, Sergei and Morozov, Stanislav and Babenko, Artem},
  journal={arXiv preprint arXiv:1909.06312},
  year={2019}
}

@inproceedings{rugamer2024scalable,
  title={Scalable Higher-Order Tensor Product Spline Models},
  author={R{\"u}gamer, David},
  booktitle={International Conference on Artificial Intelligence and Statistics},
  pages={1-9},
  year={2024},
  organization={PMLR}
}

@inproceedings{chen2016xgboost,
  title={Xgboost: A scalable tree boosting system},
  author={Chen, Tianqi and Guestrin, Carlos},
  booktitle={Proceedings of the 22nd acm sigkdd international conference on knowledge discovery and data mining},
  pages={785-794},
  year={2016}
}

@article{scikit-learn,
 title={Scikit-learn: Machine Learning in {P}ython},
 author={Pedregosa, F. and Varoquaux, G. and Gramfort, A. and Michel, V.
         and Thirion, B. and Grisel, O. and Blondel, M. and Prettenhofer, P.
         and Weiss, R. and Dubourg, V. and Vanderplas, J. and Passos, A. and
         Cournapeau, D. and Brucher, M. and Perrot, M. and Duchesnay, E.},
 journal={Journal of Machine Learning Research},
 volume={12},
 pages={2825-2830},
 year={2011}
}

@article{yahoo,
    author = {Yahoo},
    title = {Yahoo! Learning to Rank Challenge},
    journal = {https://webscope.sandbox.yahoo.com/catalog.php?datatype=c},
    year = {2010}
}

@article{micro,
  author    = {Tao Qin and
               Tie{-}Yan Liu},
  title     = {Introducing {LETOR} 4.0 Datasets},
  journal   = {CoRR},
  volume    = {abs/1306.2597},
  year      = {2013},
  url       = {http://arxiv.org/abs/1306.2597},
  bibsource = {dblp computer science bibliography, http://dblp.org}
}

@article{calhousing,
  title={Scikit-learn: Machine Learning in {P}ython},
  author={Pedregosa, F. and Varoquaux, G. and Gramfort, A. and Michel, V. and Thirion, B. and Grisel, O. and Blondel, M. and Prettenhofer, P. and Weiss, R. and Dubourg, V. and Vanderplas, J. and Passos, A. and Cournapeau, D. and Brucher, M. and Perrot, M. and Duchesnay, E.},
  journal={Journal of Machine Learning Research},
  volume={12},
  pages={2825-2830},
  year={2011}
}

@misc{wine_quality,
  author       = {Cortez,Paulo and Cerdeira,A. and Almeida,F. and Matos,T. and Reis,J.},
  title        = {{Wine Quality}},
  year         = {2009},
  howpublished = {UCI Machine Learning Repository},
  note         = {{DOI}: https://doi.org/10.24432/C56S3T}
}

@misc{drybean,
  title        = {Dry Bean},
  year         = {2020},
  howpublished = {UCI Machine Learning Repository},
  note         = {{DOI}: https://doi.org/10.24432/C50S4B}
}

@misc{Letter,
  author       = {Slate,David},
  title        = {{Letter Recognition}},
  year         = {1991},
  howpublished = {UCI Machine Learning Repository},
  note         = {{DOI}: https://doi.org/10.24432/C5ZP40}
}

@misc{credit,
  title        = {Credit Card Fraud Detection},
  year         = {2015},
  howpublished = {kaggle},
  note         = {{https://www.kaggle.com/datasets/mlg-ulb/creditcardfraud}}
}

@misc{churn,
  title        = {Telco Customer Churn},
  year         = {2017},
  howpublished = {kaggle},
  note         = {{https://www.kaggle.com/datasets/blastchar/telco-customer-churn/data}}
}

@article{wahba1995smoothing,
  title={Smoothing spline ANOVA for exponential families, with application to the Wisconsin Epidemiological Study of Diabetic Retinopathy: the 1994 Neyman Memorial Lecture},
  author={Wahba, Grace and Wang, Yuedong and Gu, Chong and Klein, Ronald and Klein, Barbara},
  journal={The Annals of Statistics},
  volume={23},
  number={6},
  pages={1865-1895},
  year={1995},
  publisher={Institute of Mathematical Statistics}
}

@inproceedings{lengerich2020purifying,
  title={Purifying interaction effects with the functional anova: An efficient algorithm for recovering identifiable additive models},
  author={Lengerich, Benjamin and Tan, Sarah and Chang, Chun-Hao and Hooker, Giles and Caruana, Rich},
  booktitle={International Conference on Artificial Intelligence and Statistics},
  pages={2402-2412},
  year={2020},
  organization={PMLR}
}

@article{liu2024kan,
  title={Kan: Kolmogorov-arnold networks},
  author={Liu, Ziming and Wang, Yixuan and Vaidya, Sachin and Ruehle, Fabian and Halverson, James and Solja{\v{c}}i{\'c}, Marin and Hou, Thomas Y and Tegmark, Max},
  journal={arXiv preprint arXiv:2404.19756},
  year={2024}
}

@book{kolmogorov1961representation,
  title={On the representation of continuous functions of several variables by superpositions of continuous functions of a smaller number of variables},
  author={Kolmogorov, Andreui Nikolaevich},
  year={1961},
  publisher={American Mathematical Society}
}

@article{hornik1989multilayer,
  title={Multilayer feedforward networks are universal approximators},
  author={Hornik, Kurt and Stinchcombe, Maxwell and White, Halbert},
  journal={Neural networks},
  volume={2},
  number={5},
  pages={359-366},
  year={1989},
  publisher={Elsevier}
}

@incollection{NEURIPS2019_9015,
title = {PyTorch: An Imperative Style, High-Performance Deep Learning Library},
author = {Paszke, Adam and Gross, Sam and Massa, Francisco and Lerer, Adam and Bradbury, James and Chanan, Gregory and Killeen, Trevor and Lin, Zeming and Gimelshein, Natalia and Antiga, Luca and Desmaison, Alban and Kopf, Andreas and Yang, Edward and DeVito, Zachary and Raison, Martin and Tejani, Alykhan and Chilamkurthy, Sasank and Steiner, Benoit and Fang, Lu and Bai, Junjie and Chintala, Soumith},
booktitle = {Advances in Neural Information Processing Systems 32},
pages = {8024-8035},
year = {2019},
publisher = {Curran Associates, Inc.},
url = {http://papers.neurips.cc/paper/9015-pytorch-an-imperative-style-high-performance-deep-learning-library.pdf}
}

@article{park2025tensor,
  title={Tensor product neural networks for functional anova model},
  author={Park, Seokhun and Kong, Insung and Choi, Yongchan and Park, Chanmoo and Kim, Yongdai},
  journal={arXiv preprint arXiv:2502.15215},
  year={2025}
}

@misc{madelon_171,
  author       = {Guyon, Isabelle},
  title        = {{Madelon}},
  year         = {2004},
  howpublished = {UCI Machine Learning Repository},
  note         = {{DOI}: https://doi.org/10.24432/C5602H}
}

@inproceedings{kishimoto2024neural,
  title={Neural Additive and Basis Models with Feature Selection and Interactions},
  author={Kishimoto, Yasutoshi and Yamanishi, Kota and Matsuda, Takuya and Shirakawa, Shinichi},
  booktitle={Pacific-Asia Conference on Knowledge Discovery and Data Mining},
  pages={3-16},
  year={2024},
  organization={Springer}
}

@article{tsang2017detecting,
  title={Detecting statistical interactions from neural network weights},
  author={Tsang, Michael and Cheng, Dehua and Liu, Yan},
  journal={arXiv preprint arXiv:1705.04977},
  year={2017}
}

@article{kohler2021rate,
  title={On the rate of convergence of fully connected deep neural network regression estimates},
  author={Kohler, Michael and Langer, Sophie},
  journal={The Annals of Statistics},
  volume={49},
  number={4},
  pages={2231-2249},
  year={2021},
  publisher={JSTOR}
}

@article{hooker_diag,
  title={Generalized functional anova diagnostics for high-dimensional functions of dependent variables},
  author={Hooker, Giles},
  journal={Journal of computational and graphical statistics},
  volume={16},
  number={3},
  pages={709-732},
  year={2007},
  publisher={Taylor \& Francis}
}

@book{gyorfi2006distribution,
  title={A distribution-free theory of nonparametric regression},
  author={Gy{\"o}rfi, L{\'a}szl{\'o} and Kohler, Michael and Krzyzak, Adam and Walk, Harro},
  year={2006},
  publisher={Springer Science \& Business Media}
}

@ARTICLE{cosso,
  author = {Lin, Yi and Zhang, Hao Helen},
  title = {COMPONENT SELECTION AND SMOOTHING IN MULTIVARIATE NONPARAMETRIC REGRESSION},
  journal = {The Annals of statistics},
  year = {2006},
  volume = {34(5)},
  pages = {2272-2297},
}

@article{hatzis2011genomic,
  title   = {A genomic predictor of response and survival following taxane-anthracycline chemotherapy for invasive breast cancer},
  author  = {Hatzis, Christos and Pusztai, Lajos and Valero, Vicente and Booser, Daniel J. and Esserman, Laura and Lluch, Ana and Vidaurre, Tatiana and Holmes, Frankie and Souchon, Eduardo and Wang, Hongkun and others},
  journal = {JAMA},
  volume  = {305},
  number  = {18},
  pages   = {1873--1881},
  year    = {2011}
}

@incollection{hastie2017generalized,
  title={Generalized additive models},
  author={Hastie, Trevor J},
  booktitle={Statistical models in S},
  pages={249--307},
  year={2017},
  publisher={Routledge}
}

@article{park2025bayesian,
  title={Bayesian Additive Regression Trees for functional ANOVA model},
  author={Park, Seokhun and Kong, Insung and Kim, Yongdai},
  journal={arXiv preprint arXiv:2509.03317},
  year={2025}
}

@article{nori2019interpretml,
  title={Interpretml: A unified framework for machine learning interpretability},
  author={Nori, Harsha and Jenkins, Samuel and Koch, Paul and Caruana, Rich},
  journal={arXiv preprint arXiv:1909.09223},
  year={2019}
}

\section{Biography Section}

\begin{IEEEbiographynophoto}{Seokhun Park}
Seokhun Park received the B.S. degree in applied mathematics from Kyung Hee University, and the Ph.D. degree in Statistics from Seoul National University.
His research interests include interpretable machine learning and functional ANOVA models.
E-mail : ajdma@naver.com
\end{IEEEbiographynophoto}

\begin{IEEEbiographynophoto}{Kunwoong Kim}
Kunwoong Kim received the B.S. degree in energy resources engineering from Seoul National University in 2019, and the Ph.D. degree in Statistics from Seoul National University.
He is currently a postdoctoral researcher in KAIST AI.
His research focuses on agentic AI, efficient inference, and trustworthy AI.
E-mail : kunwoong.kim@kaist.ac.kr
\end{IEEEbiographynophoto}

\begin{IEEEbiographynophoto}{Yongdai Kim}
Yongdai Kim received the BS and MS degrees in Statistics from Seoul National University (SNU), in 1991 and 1993, respectively, and the PhD degree of Statistics from Ohio State University, Columbus, OH, USA, in 1997. He is currently a professor with the Department of Statistics, Seoul National University. Before joining Seoul National University, he was with the National Institutes of Health (NIH), USA. His research area includes deep learning, machine learning and Bayesian statistics.
E-mail : ydkim0903@gmail.com
\end{IEEEbiographynophoto}


\clearpage
\onecolumn
\setcounter{page}{1}
\setcounter{section}{0}
\setcounter{equation}{0}
\setcounter{figure}{0}
\setcounter{table}{0}
\renewcommand{\thesection}{S.\Alph{section}}
\renewcommand{\thesectiondis}{S.\Alph{section}}

\renewcommand{\thesubsection}{\thesection.\arabic{subsection}}
\renewcommand{\thesubsectiondis}{\thesection.\arabic{subsection}}

\renewcommand{\theequation}{S\arabic{equation}}
\renewcommand{\thefigure}{S\arabic{figure}}
\renewcommand{\thetable}{S\arabic{table}}

\begin{center}
{\LARGE Supplementary Material for}\\[4pt]
{\LARGE ``SCAN: Neural Additive Networks for Scalable High-Order Component Selection and Estimation''}\\[10pt]
{\large Seokhun Park, Kunwoong Kim, and Yongdai Kim}
\end{center}
\vspace{1em}

\section{Proofs}


\subsection{Additional Notation}
Let $\mathcal{F}$ be a class of functions $f : \mathcal{X} \to \mathbb{R}$ and let $\Vert \cdot \Vert$ be a (semi-)norm on functions defined on $\mathcal{X}$.
For $\delta > 0$, a finite collection of functions $\{f_1, \ldots, f_N\}$ is called a \emph{$\delta$-cover} of $\mathcal{F}$ with respect to $\Vert \cdot \Vert$ if, for every $f \in \mathcal{F}$, there exists $j \in [N]$ such that $\Vert f - f_j \Vert \leq \delta$.
The \emph{covering number} $N(\delta, \mathcal{F}, \Vert \cdot \Vert)$ is the smallest size $N$ of a $\delta$-cover of $\mathcal{F}$ with respect to $\Vert \cdot \Vert$.
In particular, we write $N(\delta, \mathcal{F}, \Vert \cdot \Vert_{\infty})$ for the covering number with respect to the sup norm, and $N_1(\delta, \mathcal{F}, \mathbf{x}_1^n) := N(\delta, \mathcal{F}, \Vert \cdot \Vert_{1, \mathbf{x}_1^n})$ for the covering number with respect to the empirical $L^1$ norm on $\mathbf{x}_1^n$.
For two positive sequences (or functions) $a_n$ and $b_n$, we write $a_n \lesssim b_n$ if there exists a constant $C > 0$, independent of $n$, such that $a_n \leq C\, b_n$ for all sufficiently large $n$.

\subsection{Properties of the Identifiable Projection}\label{app:projection}

\begin{proposition} Assume that $\mathcal{M}_{t}$ depends only on $\mathbf{x}_{S_{t}}$.
For $S \not \subseteq S_{t}$, we have
$\Pi_{S}\mathcal{M}_{t} = 0$.
\label{prop:projection-0}
\end{proposition}
\noindent\textbf{Proof.}
We have
$\mathcal{M}_t(\mathbf{x}) = h\big(\sum_{j \in S_t} \tilde\alpha_{j,t}\,\mathcal{B}_{j,t}(x_j)\big)$,
so $\mathcal{M}_t$ depends on $\mathbf{x}$ only through $\mathbf{x}_{S_t}$.\\
$\newline$
Since $S \not\subseteq S_t$, there exists $k \in S \setminus S_t$.
We first show that, for every $U \subseteq S \setminus \{k\}$,
\begin{align}
(\mathbb{E}_\nu \mathcal{M}_t)(\mathbf{x}_{U \cup \{k\}}) = (\mathbb{E}_\nu \mathcal{M}_t)(\mathbf{x}_U).
\label{eq:drop_k}
\end{align}
Write $U' := U \cup \{k\}$.
Since $\mathcal{M}_t$ does not depend on its $k$-th argument, we have
\begin{align*}
(\mathbb{E}_\nu \mathcal{M}_t)(\mathbf{x}_U)
&= \int \!\! \int \mathcal{M}_t(\mathbf{x}_U, z_k, \mathbf{z}_{-U'})\, \nu_k(dz_k)\, \nu_{-U'}(d\mathbf{z}_{-U'}) \\
&= \int \mathcal{M}_t(\mathbf{x}_U, x_k, \mathbf{z}_{-U'}) \Big( \int \nu_k(dz_k) \Big)\, \nu_{-U'}(d\mathbf{z}_{-U'}) \\
&= \int \mathcal{M}_t(\mathbf{x}_{U'}, \mathbf{z}_{-U'})\, \nu_{-U'}(d\mathbf{z}_{-U'})
= (\mathbb{E}_\nu \mathcal{M}_t)(\mathbf{x}_{U'}),
\end{align*}
which proves (\ref{eq:drop_k}).\\
$\newline$
Using (\ref{eq:purify}) and (\ref{eq:drop_k}), we have 
\begin{align*}
\Pi_S \mathcal{M}_t(\mathbf{x})
&= \sum_{U \subseteq S} (-1)^{|S| - |U|} (\mathbb{E}_{\nu} \mathcal{M}_{t})(\mathbf{x}_U)\\
&= \sum_{U \subseteq S \backslash \{k\}} (-1)^{|S| - |U|} (\mathbb{E}_{\nu} \mathcal{M}_{t})(\mathbf{x}_U)
 + \sum_{U \subseteq S,\, k \in U} (-1)^{|S| - |U|} (\mathbb{E}_{\nu} \mathcal{M}_{t})(\mathbf{x}_U)\\
&= \sum_{U \subseteq S \setminus \{k\}} (-1)^{|S| - |U|}
\Big\{ (\mathbb{E}_\nu \mathcal{M}_t)(\mathbf{x}_U) - (\mathbb{E}_\nu \mathcal{M}_t)(\mathbf{x}_{U \cup \{k\}}) \Big\} = 0.
\end{align*}
\qed

\begin{proposition}\label{prop:identifiability}
Let $g$ be a $\nu$-integrable function on $\mathcal{X}$ and $S \subseteq [p]$ be nonempty.
Then, for every $j \in S$,
\begin{align*}
\int_{\mathcal{X}_j} \Pi_S g(\mathbf{x}_S)\, \nu_j(dx_j) = 0.
\end{align*}
\end{proposition}

\begin{proof}
Fix $j \in S$.
Splitting the subsets of $S$ according to whether they contain $j$, we can write (\ref{eq:purify}) as
\begin{align*}
\Pi_S g(\mathbf{x}) = \sum_{U \subseteq S \setminus \{j\}} (-1)^{|S|-|U|}
\Big\{ (\mathbb{E}_\nu g)(\mathbf{x}_U) - (\mathbb{E}_\nu g)(\mathbf{x}_{U \cup \{j\}}) \Big\}.
\end{align*}
Since $(\mathbb{E}_\nu g)(\mathbf{x}_U)$ does not depend on $x_j$, we have
$\int (\mathbb{E}_\nu g)(\mathbf{x}_U)\, \nu_j(dx_j) = (\mathbb{E}_\nu g)(\mathbf{x}_U)$.
Moreover,
\begin{align*}
\int (\mathbb{E}_\nu g)(\mathbf{x}_{U \cup \{j\}})\, \nu_j(dx_j)
&= \int \!\! \int g(\mathbf{x}_U, x_j, \mathbf{z}_{-(U \cup \{j\})})\,
   \nu_{-(U \cup \{j\})}(d\mathbf{z}_{-(U \cup \{j\})})\, \nu_j(dx_j) \\
&= (\mathbb{E}_\nu g)(\mathbf{x}_U).
\end{align*}
Hence each term in the braces integrates to zero with respect to $\nu_j$, and the claim follows.
\end{proof}

\subsection{Proof of Theorem \ref{thm:approx}}

Without loss of generality, let $S = \{1, \ldots, d\}$ and $r = T^{1/d} \in \mathbb{N}$.
For $\mathbf{k} \in \mathbb{Z}^d$ and $\boldsymbol{\eta} \in \{0,1\}^d$, write $|\boldsymbol{\eta}| := \sum_{i} \eta_i$,
and let $\mathbf{e}_i$ denote the $i$-th standard basis vector of $\mathbb{R}^d$.
The proof proceeds as follows.
In Step 1, we define a grid and the output weights.
In Step 2, we construct the basis functions.
In Step 3, we describe the behavior of each basis function away from its thresholds.
In Step 4, we rewrite the SCAN function by summation by parts, and in Step 5, we localize it around each $\mathbf{x}$.
Step 6 shows that truncation does not change the function, and Step 7 combines the errors.
Throughout, we use $\sigma(-u) \leq e^{-u}$ for $u \geq 0$ and $|\sigma(u) - \sigma(u')| \leq |u - u'|/4$.
\\
$\newline$
\textbf{Step 1: Grid and output weights.}
Define the thresholds
\begin{align*}
t_0 := -\frac{1}{r}, \qquad t_l := \frac{l}{r} \quad (l = 1, \ldots, r),
\end{align*}
and, for $\mathbf{m} \in [r]^d$, the grid point $\mathbf{z}_{\mathbf{m}} := \big( (m_1 - 1)/r, \ldots, (m_d - 1)/r \big) \in [0,1)^d$ and
$c_{\mathbf{m}} := f_{0,S}(\mathbf{z}_{\mathbf{m}})$.
Set $c_{\mathbf{m}} := 0$ if $m_i = 0$ for some $i$, and define
\begin{align*}
w_{\mathbf{k}} := \sum_{\boldsymbol{\eta} \in \{0,1\}^{d}} (-1)^{|\boldsymbol{\eta}|}\, c_{\mathbf{k} - \boldsymbol{\eta}},
\qquad \mathbf{k} \in [r]^d.
\end{align*}
Since $w_{\mathbf{k}}$ is a signed sum of $2^d$ values bounded by $F$ in absolute value, $|w_{\mathbf{k}}| \leq 2^d F$.
\\
$\newline$
\textbf{Step 2: Construction of the basis functions.}
We assign one basis function $\mathcal{M}_{\mathbf{k}}$ with output weight $w_{\mathbf{k}}$ to each $\mathbf{k} \in [r]^{d}$, so that $T = r^{d}$.
Let
\begin{align*}
\delta := \frac{1}{4r}, \qquad \gamma := \frac{\delta}{10dr} = \frac{1}{40d\, r^{2}}.
\end{align*}

\textit{Component-selection weights.}
Let $\theta_{i,\mathbf{k}} = B_\theta$ for $i \in S$ and $\theta_{j,\mathbf{k}} = -B_\theta$ for $j \notin S$, and write $q := e^{-2B_\theta/\mathbb{T}} \in (0,1]$. Then
\begin{align*}
\alpha_{i,\mathbf{k}} = \frac{1}{d + (p-d)q} \quad (i \in S), \qquad
\alpha_{j,\mathbf{k}} = \frac{q}{d + (p-d)q} \leq \frac{q}{d} \quad (j \notin S).
\end{align*}
Since $0 < \epsilon_\alpha < 1/d$, we can choose
\begin{align*}
B_\theta > \frac{\mathbb{T}}{2} \log \max\Big\{ \frac{1}{d\epsilon_\alpha},\; \frac{(p-d)\epsilon_\alpha}{1 - d\epsilon_\alpha} \Big\}
\end{align*}
(the second term is absent if $p = d$), so that $q \leq d\epsilon_\alpha$ and $q < (1 - d\epsilon_\alpha)/\{(p-d)\epsilon_\alpha\}$.
Then $\alpha_{j,\mathbf{k}} \leq \epsilon_\alpha$ for $j \notin S$ and $\alpha_{i,\mathbf{k}} > \epsilon_\alpha$ for $i \in S$, so that $S_{\mathbf{k}} = S$.
Moreover, $\alpha_{i,\mathbf{k}} \geq 1/p$ for $i \in S$ since $q \leq 1$, and $\alpha_{i,\mathbf{k}}$ does not depend on $\mathbf{k}$.

\textit{Univariate unit functions.}
For $i \in S$, let $\beta_{i,\mathbf{k},1} = -2r / \alpha_{i,\mathbf{k}}$, $\beta_{i,\mathbf{k},2} = r / (d\, \alpha_{i,\mathbf{k}})$, $b_{i,\mathbf{k}} = t_{k_i-1}$, and $\gamma_{i,\mathbf{k}} = \gamma$.
Then $\alpha_{i,\mathbf{k}} \mathcal{B}_{i,\mathbf{k}}(x_i) = \phi(x_i - t_{k_i - 1})$, where
\begin{align*}
\phi(u) := -2r \Big( 1 - \sigma\Big( \frac{u}{\gamma} \Big) \Big) + \frac{r}{d}\, \sigma\Big( \frac{u}{\gamma} \Big).
\end{align*}
For $j \notin S$, let $\beta_{j,\mathbf{k},1} = \beta_{j,\mathbf{k},2} = 0$, $b_{j,\mathbf{k}} = 0$, and $\gamma_{j,\mathbf{k}} = 1$, so that $\mathcal{B}_{j,\mathbf{k}} \equiv 0$.
Then,
\begin{align*}
\mathcal{M}_{\mathbf{k}}(\mathbf{x}) = \sigma\bigg( \sum_{i=1}^{d} \phi(x_i - t_{k_i-1}) \bigg).
\end{align*}
For convenience, we set $\mathcal{M}_{\mathbf{k}} :\equiv 0$ if $k_i = r+1$ for some $i$.

\textit{Parameter bound.}
Since $|w_{\mathbf{k}}| \leq 2^d F$, $|\beta_{i,\mathbf{k},1}|, |\beta_{i,\mathbf{k},2}| \leq 2rp$, $|b_{i,\mathbf{k}}| \leq 1$, and $\gamma \leq 1$, $1/\gamma = 40d\, r^{2}$,
the constructed parameters belong to $\Psi(T, B)$ whenever
\begin{align*}
B \geq \max\Big\{ 2^d F,\; 2p\, T^{1/d},\; 40d\, T^{2/d},\; B_\theta \Big\}.
\end{align*}
\\
$\newline$
\textbf{Step 3: Behavior away from the thresholds.}
If $|u| \geq \delta$, then $|u|/\gamma \geq 10dr$, so that $\sigma(u/\gamma)$ is within $e^{-10dr}$ of $0$ or $1$.
Using $3r e^{-10dr} \leq r/(4d)$, which holds since $12d\, e^{-10d} \leq 1$, we obtain
\begin{align}
u \geq \delta \;&\Rightarrow\; \phi(u) \in \Big[ \frac{r}{d} - 3r e^{-10dr},\, \frac{r}{d} \Big) \subseteq \Big[\frac{3r}{4d},\, \frac{r}{d}\Big) \quad \text{(pass)}, \label{eq:pass}\\
u \leq -\delta \;&\Rightarrow\; \phi(u) \leq -2r + 3r e^{-10dr} \leq -2r + \frac{r}{4d} \quad \text{(fail)}. \nonumber
\end{align}
Moreover, $-2r < \phi(u) < r/d$ for all $u$.
Hence, for $\mathbf{k} \in [r]^d$ and $\mathbf{x} \in \mathcal{X}$:
\begin{enumerate}
\item[(a)] If $x_i \leq t_{k_i-1} - \delta$ for some $i \in S$, then $\sum_{i} \phi(x_i - t_{k_i-1}) \leq (d-1)\frac{r}{d} - 2r + \frac{r}{4d} \leq -\frac{r}{2}$, and thus $0 \leq \mathcal{M}_{\mathbf{k}}(\mathbf{x}) \leq e^{-r/2}$.
\item[(b)] If $x_i \geq t_{k_i-1} + \delta$ for all $i \in S$, then $\sum_{i} \phi(x_i - t_{k_i-1}) \geq \frac{3r}{4} \geq \frac{r}{2}$, and thus $1 - e^{-r/2} \leq \mathcal{M}_{\mathbf{k}}(\mathbf{x}) \leq 1$.
\item[(c)] If $\mathbf{k} + \mathbf{e}_i \in [r]^d$ and $x_i \geq t_{k_i} + \delta$, then $x_i - t_{k_i-1} \geq \delta$ and $x_i - t_{k_i} \geq \delta$.
By (\ref{eq:pass}), $|\phi(x_i - t_{k_i-1}) - \phi(x_i - t_{k_i})| \leq 3r e^{-10dr}$.
Since $\mathcal{M}_{\mathbf{k}}$ and $\mathcal{M}_{\mathbf{k}+\mathbf{e}_i}$ differ only in the $i$-th summand,
\begin{align*}
|\mathcal{M}_{\mathbf{k}}(\mathbf{x}) - \mathcal{M}_{\mathbf{k}+\mathbf{e}_i}(\mathbf{x})| \leq \frac{3r}{4} e^{-10dr} \leq e^{-r/2}.
\end{align*}
\end{enumerate}
$\newline$
\textbf{Step 4: Summation by parts.}
By the definition of $w_{\mathbf{k}}$ and the substitution $\mathbf{m} = \mathbf{k} - \boldsymbol{\eta}$, using $c_{\mathbf{m}} = 0$ if $m_i = 0$ for some $i$ and $\mathcal{M}_{\mathbf{k}} \equiv 0$ if $k_i = r+1$ for some $i$,
\begin{align}
\begin{split}
&f_{\text{SCAN},S}(\mathbf{x} \mid \boldsymbol{\psi}, 0)
= \sum_{\mathbf{k} \in [r]^d} w_{\mathbf{k}} \mathcal{M}_{\mathbf{k}}(\mathbf{x})
= \sum_{\mathbf{m} \in [r]^d} c_{\mathbf{m}} D_{\mathbf{m}}(\mathbf{x}),\\
&D_{\mathbf{m}}(\mathbf{x}) := \sum_{\boldsymbol{\eta} \in \{0,1\}^d} (-1)^{|\boldsymbol{\eta}|} \mathcal{M}_{\mathbf{m} + \boldsymbol{\eta}}(\mathbf{x}).
\end{split}
\label{eq:sbp}
\end{align}
That is, $D_{\mathbf{m}} = (\Delta_1 \cdots \Delta_d \mathcal{M})_{\mathbf{m}}$ with $(\Delta_i a)_{\mathbf{m}} := a_{\mathbf{m}} - a_{\mathbf{m} + \mathbf{e}_i}$.
Since $\sum_{m_i = 1}^{r} (\Delta_i a)_{\mathbf{m}} = a_{\mathbf{m}}|_{m_i = 1} - a_{\mathbf{m}}|_{m_i = r+1}$ and $\mathcal{M}_{\mathbf{k}} \equiv 0$ if $k_i = r+1$, summing over each coordinate in turn gives
\begin{align*}
\sum_{\mathbf{m} \in [r]^d} D_{\mathbf{m}}(\mathbf{x}) = \mathcal{M}_{\mathbf{1}}(\mathbf{x}), \qquad \mathbf{1} := (1, \ldots, 1).
\end{align*}
The thresholds of $\mathcal{M}_{\mathbf{1}}$ are all $t_0 = -1/r$, and $x_i \geq 0 \geq t_0 + \delta$ for every $\mathbf{x} \in \mathcal{X}$.
Hence, by (b),
\begin{align}
\Big| \sum_{\mathbf{m} \in [r]^d} D_{\mathbf{m}}(\mathbf{x}) - 1 \Big| \leq e^{-r/2}.
\label{eq:sum_D}
\end{align}
Moreover, since $D_{\mathbf{m}}(\mathbf{x})$ is the difference of two sums of $2^{d-1}$ values in $[0,1]$,
\begin{align}
|D_{\mathbf{m}}(\mathbf{x})| \leq 2^{d-1}.
\label{eq:D_trivial}
\end{align}
\\
$\newline$
\textbf{Step 5: Localization.}
Fix $\mathbf{x} \in \mathcal{X}$ and let
\begin{align*}
\mathcal{K}(\mathbf{x}) := \big\{ \mathbf{m} \in [r]^d :\; t_{m_i - 1} - \delta < x_i \ \text{and}\ \big( m_i = r \ \text{or}\ x_i < t_{m_i} + \delta \big) \ \text{for all } i \in S \big\}.
\end{align*}

\textit{(i) Away from $\mathbf{x}$.}
Let $\mathbf{m} \notin \mathcal{K}(\mathbf{x})$. Then one of the following holds.
\begin{itemize}
\item There exists $i$ with $x_i \leq t_{m_i-1} - \delta$.
For every $\boldsymbol{\eta}$, either $\mathcal{M}_{\mathbf{m}+\boldsymbol{\eta}} \equiv 0$, or its $i$-th threshold satisfies $t_{m_i - 1 + \eta_i} \geq t_{m_i-1}$, so that $0 \leq \mathcal{M}_{\mathbf{m}+\boldsymbol{\eta}}(\mathbf{x}) \leq e^{-r/2}$ by (a).
Hence $|D_{\mathbf{m}}(\mathbf{x})| \leq 2^{d-1} e^{-r/2}$.
\item There exists $i$ with $m_i \leq r-1$ and $x_i \geq t_{m_i} + \delta$.
Pairing $\boldsymbol{\eta}$ with $\eta_i = 0$ and $\boldsymbol{\eta} + \mathbf{e}_i$,
\begin{align*}
D_{\mathbf{m}}(\mathbf{x}) = \sum_{\boldsymbol{\eta} : \eta_i = 0} (-1)^{|\boldsymbol{\eta}|} \big( \mathcal{M}_{\mathbf{m}+\boldsymbol{\eta}}(\mathbf{x}) - \mathcal{M}_{\mathbf{m}+\boldsymbol{\eta}+\mathbf{e}_i}(\mathbf{x}) \big).
\end{align*}
If $\mathbf{m} + \boldsymbol{\eta} \notin [r]^d$, both terms vanish.
Otherwise, $\mathbf{m}+\boldsymbol{\eta}+\mathbf{e}_i \in [r]^d$ since $m_i \leq r-1$, and (c) applies.
Hence $|D_{\mathbf{m}}(\mathbf{x})| \leq 2^{d-1} e^{-r/2}$.
\end{itemize}

\textit{(ii) Size of $\mathcal{K}(\mathbf{x})$.}
Fix $i$, and suppose that $m_i$ and $m'_i \geq m_i + 2$ both satisfy the condition in $\mathcal{K}(\mathbf{x})$.
Then $x_i < t_{m_i} + \delta$ and $x_i > t_{m'_i - 1} - \delta \geq t_{m_i + 1} - \delta$, so that $1/r = t_{m_i+1} - t_{m_i} < 2\delta = 1/(2r)$, which is a contradiction.
Hence, for each $i$, at most two consecutive values of $m_i$ are allowed, and $|\mathcal{K}(\mathbf{x})| \leq 2^d$.

\textit{(iii) Near $\mathbf{x}$.}
Let $\mathbf{m} \in \mathcal{K}(\mathbf{x})$ and $i \in S$.
If $m_i \geq 2$, then $z_{\mathbf{m},i} = t_{m_i-1}$ and $x_i \in (t_{m_i-1} - \delta,\, t_{m_i-1} + 1/r + \delta)$ (for $m_i = r$, we use $x_i \leq 1 = t_{r-1} + 1/r$).
If $m_i = 1$, then $z_{\mathbf{m},i} = 0$ and $x_i \in [0,\, 1/r + \delta)$ (for $r = 1$, we use $x_i \leq 1$).
In both cases, $|x_i - z_{\mathbf{m},i}| \leq 1/r + \delta = 5/(4r)$.
Hence, by assumption (i),
\begin{align*}
|c_{\mathbf{m}} - f_{0,S}(\mathbf{x}_S)|
= |f_{0,S}(\mathbf{z}_{\mathbf{m}}) - f_{0,S}(\mathbf{x}_S)|
\leq L \Big( \frac{5\sqrt{d}}{4r} \Big)^{\rho}
\leq \frac{5}{4} \cdot \frac{L d^{\rho/2}}{r^{\rho}}.
\end{align*}
\\
$\newline$
\textbf{Step 6: Truncation does not change the function.}
By Step 2, $\alpha_{j,\mathbf{k}} \leq \epsilon_\alpha < \alpha_{i,\mathbf{k}}$ for all $j \notin S$ and $i \in S$, so the truncation sets $\tilde{\alpha}_{j,\mathbf{k}} = 0$ for $j \notin S$ and keeps $\tilde{\alpha}_{i,\mathbf{k}} = \alpha_{i,\mathbf{k}}$ for $i \in S$.
Since $\mathcal{B}_{j,\mathbf{k}} \equiv 0$ for $j \notin S$, the input of each activation function is unchanged, and hence
\begin{align*}
f_{\text{SCAN},S}(\cdot \mid \boldsymbol{\psi},\epsilon_{\alpha}) = f_{\text{SCAN},S}(\cdot \mid \boldsymbol{\psi},0).
\end{align*}
\\
$\newline$
\textbf{Step 7: Combining the errors.}
By Step 6, (\ref{eq:sbp}), and
\begin{align*}
&f_{\text{SCAN},S}(\mathbf{x} \mid \boldsymbol{\psi},0) - f_{0,S}(\mathbf{x}_S)
\\
&= \sum_{\mathbf{m} \in [r]^d} \big( c_{\mathbf{m}} - f_{0,S}(\mathbf{x}_S) \big) D_{\mathbf{m}}(\mathbf{x})
+ f_{0,S}(\mathbf{x}_S) \Big( \sum_{\mathbf{m} \in [r]^d} D_{\mathbf{m}}(\mathbf{x}) - 1 \Big),
\end{align*}
we split the first sum over $\mathcal{K}(\mathbf{x})$ and its complement.
Using Step 5 (ii)-(iii) and (\ref{eq:D_trivial}) on $\mathcal{K}(\mathbf{x})$,
Step 5 (i) and $|c_{\mathbf{m}} - f_{0,S}(\mathbf{x}_S)| \leq 2F$ on its complement, and (\ref{eq:sum_D}) for the last term, we obtain
\begin{align*}
\big| f_{0,S}(\mathbf{x}_S) - f_{\text{SCAN},S}(\mathbf{x} \mid \boldsymbol{\psi},\epsilon_{\alpha}) \big|
&\leq 2^{d} \cdot 2^{d-1} \cdot \frac{5}{4} \cdot \frac{L d^{\rho/2}}{r^{\rho}}
+ r^d \cdot 2F \cdot 2^{d-1} e^{-r/2} + F e^{-r/2} \\
&\leq \frac{4^{d} L d^{\rho/2}}{r^{\rho}} + 2^{d+1} F r^{d} e^{-r/2}.
\end{align*}
Since this holds for every $\mathbf{x} \in \mathcal{X}$ and $r = T^{1/d}$, this proves Theorem~\ref{thm:approx}.

\qed

\newpage

\subsection{Covering Number}

We consider the class of truncated SCAN functions with bounded parameters,
\begin{align*}
\mathcal{F}(T, B, \epsilon_{\alpha})
 := \big\{ f_{\text{SCAN}}(\cdot \mid \boldsymbol{\psi}, \epsilon_{\alpha}) : \boldsymbol{\psi} \in \Psi(T, B) \big\},
\end{align*}
where the temperature $\mathbb{T} > 0$ is fixed and $(\alpha_{1,t}, \ldots, \alpha_{p,t})^\top = \text{softmax}(\boldsymbol{\Theta}_t / \mathbb{T})$.
Note that the number of parameters is $P := T(5p + 1)$.

\begin{lemma}[Covering number]\label{lem:covering}
Let $\mathcal{X} = [0,1]^{p}$. For any $\delta > 0$ and $\epsilon_{\alpha} > 0$,
\begin{align*}
\log N\big(\delta, \mathcal{F}(T, B,\epsilon_{\alpha}), \Vert \cdot \Vert_{\infty}\big)
\leq pT \log 2 + T(5p + 1) \log\left( 1 + \frac{T(5p+1) B^{7}}{\delta} \right).
\end{align*}
\end{lemma}

The difficulty is that the truncated component-selection weights $\tilde\alpha_{j,t} = \alpha_{j,t}\, \mathbb{I}(\alpha_{j,t} > \epsilon_{\alpha})$ are discontinuous in $\boldsymbol{\Theta}_t$, and hence $f_{\text{SCAN}}(\cdot \mid \boldsymbol{\psi}, \epsilon_{\alpha})$ is not Lipschitz continuous in $\boldsymbol{\psi}$.
We overcome this by fixing the set of features selected by each basis function.
The proof consists of three steps.
In Step 1, we decompose $\mathcal{F}(T, B, \epsilon_\alpha)$ according to the selected features.
In Step 2, we show that each piece is Lipschitz continuous in its parameters.
In Step 3, we cover each piece by a grid and take the union.
Throughout, we use $|\sigma(u)| \leq 1$ and $|\sigma(u) - \sigma(u')| \leq |u - u'|/4$.
\\
$\newline$
\textbf{Step 1: Decomposition by the selected features.}
For $\mathbf{S} = (S_1, \ldots, S_T) \in \mathcal{P}([p])^{T}$ and $\boldsymbol{\psi} \in \Psi(T, B)$, define
\begin{align*}
f_{\mathbf{S}}(\mathbf{x} \mid \boldsymbol{\psi}) := \sum_{t=1}^{T} w_t\, \sigma\bigg( \sum_{j \in S_t} \alpha_{j,t} \mathcal{B}_{j,t}(x_j) \bigg)
\end{align*}
and
\begin{align*}
\mathcal{F}_{\mathbf{S}} := \big\{ f_{\mathbf{S}}(\cdot \mid \boldsymbol{\psi}) : \boldsymbol{\psi} \in \Psi(T, B) \big\}.
\end{align*}
For each $\boldsymbol{\psi} \in \Psi(T, B)$, let $S_t(\boldsymbol{\psi}) := \{ j \in [p] : \alpha_{j,t} > \epsilon_{\alpha} \}$.
Then $f_{\text{SCAN}}(\cdot \mid \boldsymbol{\psi}, \epsilon_{\alpha}) = f_{\mathbf{S}(\boldsymbol{\psi})}(\cdot \mid \boldsymbol{\psi})$ with $\mathbf{S}(\boldsymbol{\psi}) := (S_1(\boldsymbol{\psi}), \ldots, S_T(\boldsymbol{\psi}))$, and hence
\begin{align}
\mathcal{F}(T, B,\epsilon_{\alpha}) \subseteq \bigcup_{\mathbf{S} \in \mathcal{P}([p])^{T}} \mathcal{F}_{\mathbf{S}}.
\label{eq:cover_union}
\end{align}
Note that the number of $\mathbf{S}$ is $|\mathcal{P}([p])^{T}| = 2^{pT}$.
\\
$\newline$
\textbf{Step 2: Lipschitz continuity in the parameters.}
Fix $\mathbf{S} \in \mathcal{P}([p])^{T}$.
Let $\boldsymbol{\psi}, \boldsymbol{\psi}' \in \Psi(T, B)$. We denote the quantities computed with $\boldsymbol{\psi}'$ by primes, and write $\Delta a := a - a'$ for any quantity $a$. For $\mathbf{x} \in [0,1]^{p}$, let
\begin{align*}
u_t := \sum_{j \in S_t} \alpha_{j,t} \mathcal{B}_{j,t}(x_j), \qquad
s_{j,t} := \sigma\Big( \frac{x_j - b_{j,t}}{\gamma_{j,t}} \Big),
\end{align*}
so that $f_{\mathbf{S}}(\mathbf{x} \mid \boldsymbol{\psi}) = \sum_{t=1}^{T} w_t \sigma(u_t)$ and $\mathcal{B}_{j,t}(x_j) = \beta_{j,t,1}(1 - s_{j,t}) + \beta_{j,t,2} s_{j,t}$.
Since the index sets $S_t$ are fixed, $u_t$ depends on $\boldsymbol{\Theta}_t$ only through the continuous map $\boldsymbol{\Theta}_t \mapsto (\alpha_{j,t})_{j \in S_t}$.

\textit{(a) Univariate unit functions.}
Since $\mathcal{B}_{j,t}(x_j)$ is a convex combination of $\beta_{j,t,1}$ and $\beta_{j,t,2}$, $|\mathcal{B}_{j,t}(x_j)| \leq B$.
Moreover, with $z = (x_j - b_{j,t})/\gamma_{j,t}$, we have
$|\partial \mathcal{B}_{j,t}/\partial \beta_{j,t,1}|, |\partial \mathcal{B}_{j,t}/\partial \beta_{j,t,2}| \leq 1$,
$|\partial \mathcal{B}_{j,t}/\partial b_{j,t}| = |\beta_{j,t,2} - \beta_{j,t,1}|\, \sigma'(z)/\gamma_{j,t} \leq B^{2}/2$, and
$|\partial \mathcal{B}_{j,t}/\partial \gamma_{j,t}| = |\beta_{j,t,2} - \beta_{j,t,1}|\, \sigma'(z)\, |x_j - b_{j,t}|/\gamma_{j,t}^{2} \leq B^{3}(1+B)/2 \leq B^{4}$,
using $\sigma' \leq 1/4$, $|x_j - b_{j,t}| \leq 1 + B$, and $\gamma_{j,t} \geq 1/B$.
Since the parameter set is a box, the mean value theorem along the segment between the parameters gives
\begin{align*}
|\Delta \mathcal{B}_{j,t}|
\leq B^{4} \big( |\Delta\beta_{j,t,1}| + |\Delta\beta_{j,t,2}| + |\Delta b_{j,t}| + |\Delta\gamma_{j,t}| \big).
\end{align*}

\textit{(b) Component-selection weights.}
Since $\Vert \text{softmax}(\mathbf{z}) - \text{softmax}(\mathbf{z}') \Vert_{1} \leq 2 \Vert \mathbf{z} - \mathbf{z}' \Vert_{\infty}$ and $1/\mathbb{T} \leq B$,
\begin{align*}
\sum_{j \in S_t} |\Delta \alpha_{j,t}|
\leq \sum_{j=1}^{p} |\Delta \alpha_{j,t}|
\leq \frac{2}{\mathbb{T}} \Vert \Delta \boldsymbol{\Theta}_t \Vert_{\infty}
\leq 2B \Vert \Delta \boldsymbol{\Theta}_t \Vert_{1}.
\end{align*}

\textit{(c) Input of the activation function.}
Since $\sum_{j \in S_t} \alpha'_{j,t} \leq 1$, we have $|u'_t| \leq B$, and by (a) and (b),
\begin{align*}
|\Delta u_t|
&\leq \sum_{j \in S_t} \big( |\Delta\alpha_{j,t}| |\mathcal{B}_{j,t}| + \alpha'_{j,t} |\Delta\mathcal{B}_{j,t}| \big) \\
&\leq 2B^{2} \Vert \Delta\boldsymbol{\Theta}_t \Vert_{1} + B^{4} \sum_{j=1}^{p} \big( |\Delta\beta_{j,t,1}| + |\Delta\beta_{j,t,2}| + |\Delta b_{j,t}| + |\Delta\gamma_{j,t}| \big).
\end{align*}
Since $2B^2 \leq B^4$ for $B \geq 2$, it follows that
\begin{align*}
|\Delta u_t| \leq B^{4} \Vert \Delta \boldsymbol{\zeta}_t \Vert_{1} \leq B^{5} \Vert \Delta \boldsymbol{\zeta}_t \Vert_{1}.
\end{align*}

\textit{(d) Output.}
Finally,
\begin{align*}
\big| f_{\mathbf{S}}(\mathbf{x} \mid \boldsymbol{\psi}) - f_{\mathbf{S}}(\mathbf{x} \mid \boldsymbol{\psi}') \big|
\leq \sum_{t=1}^{T} \Big( |\Delta w_t| + |w'_t| \cdot \frac{1}{4} |\Delta u_t| \Big)
\leq B^{6} \Vert \boldsymbol{\psi} - \boldsymbol{\psi}' \Vert_{1}.
\end{align*}
Since this holds for all $\mathbf{x} \in [0,1]^{p}$,
\begin{align}
\Vert f_{\mathbf{S}}(\cdot \mid \boldsymbol{\psi}) - f_{\mathbf{S}}(\cdot \mid \boldsymbol{\psi}') \Vert_{\infty} \leq B^{6} \Vert \boldsymbol{\psi} - \boldsymbol{\psi}' \Vert_{1}.
\label{eq:lipschitz}
\end{align}
\\
$\newline$
\textbf{Step 3: Covering the parameter space.}
Every coordinate of $\boldsymbol{\psi} \in \Psi(T, B)$ lies in an interval of length at most $2B$, and $\boldsymbol{\psi}$ has $P = T(5p+1)$ coordinates.
Let $\eta = \delta / (P B^{6})$. Taking a grid of radius $\eta$ in each coordinate, we obtain a set $\{\boldsymbol{\psi}^{(1)}, \ldots, \boldsymbol{\psi}^{(N_0)}\} \subseteq \Psi(T,B)$ with
\begin{align*}
N_0 \leq \left( 1 + \frac{B}{\eta} \right)^{P} = \left( 1 + \frac{P B^{7}}{\delta} \right)^{P}
\end{align*}
such that, for every $\boldsymbol{\psi} \in \Psi(T, B)$, there exists $\boldsymbol{\psi}^{(i)}$ with $\Vert \boldsymbol{\psi} - \boldsymbol{\psi}^{(i)} \Vert_{1} \leq P\eta = \delta / B^{6}$.
By (\ref{eq:lipschitz}), $\{ f_{\mathbf{S}}(\cdot \mid \boldsymbol{\psi}^{(i)}) \}_{i=1}^{N_0}$ is a $\delta$-cover of $\mathcal{F}_{\mathbf{S}}$ in $\Vert \cdot \Vert_{\infty}$ for each $\mathbf{S}$.
Taking the union of these covers over all $\mathbf{S} \in \mathcal{P}([p])^{T}$ and using (\ref{eq:cover_union}), we obtain a $\delta$-cover of $\mathcal{F}(T, B, \epsilon_{\alpha})$ of size at most $2^{pT} N_0$, which completes the proof.

\qed


\newpage

\subsection{Proof of Theorem \ref{thm:conv_rate}}

We write $m := f_0$ and
\begin{align*}
\mathcal{F}_n := \mathcal{F}(T_n, B_n, \epsilon_\alpha),
\end{align*}
so that $\hat f^{\mathrm{raw}}_n \in \operatorname*{argmin}_{f \in \mathcal{F}_n} n^{-1} \sum_{i=1}^n (Y_i - f(\mathbf{X}_i))^2$ and $\hat f_n = \mathcal{T}_{\beta_n} \hat f^{\mathrm{raw}}_n$.
We use the following oracle inequality for truncated least squares estimators.

\begin{lemma}\label{lem:oracle}
Assume that $\mathbb{E}\{\exp(c_0 Y^2)\} < \infty$ and $\Vert m \Vert_\infty < \infty$.
Let $\mathcal{F}_n$ be a class of functions, $\beta_n = c_1 \log n$,
$\hat m_n \in \operatorname*{argmin}_{f \in \mathcal{F}_n} n^{-1} \sum_{i=1}^n (Y_i - f(\mathbf{X}_i))^2$,
and $\mathcal{T}_{\beta_n}\mathcal{F}_n := \{\mathcal{T}_{\beta_n} f : f \in \mathcal{F}_n\}$. Then, for $n > 1$,
\begin{align*}
&\mathbb{E} \int |\mathcal{T}_{\beta_n}\hat{m}_n - m|^{2} d\mu \\
&\leq \frac{c_3 (\log n)^{2} \sup_{\mathbf{x}_1^n} \big( \log N_1\big( \frac{1}{n \beta_n}, \mathcal{T}_{\beta_n}\mathcal{F}_n, \mathbf{x}_1^n \big) + 1 \big)}{n}
+ 2 \inf_{f \in \mathcal{F}_n} \int |f - m|^{2} d\mu .
\end{align*}
\end{lemma}
\textbf{Proof.} This is Lemma 18 in Supplement B of \cite{kohler2021rate}. \qed
$\newline$
$\newline$
To prove Theorem~\ref{thm:conv_rate}, we bound the two terms in Lemma~\ref{lem:oracle}.
Note that $\Vert m \Vert_{\infty} \leq |\mathcal{S}| F < \infty$.
Let $k_n := \lceil n^{1/(2\rho + d)} \rceil$, so that $n^{1/(2\rho+d)} \leq k_n \leq 2 n^{1/(2\rho+d)}$.
\\
$\newline$
\textbf{Step 1: Approximation term.}
Fix $S \in \mathcal{S}$. Since $\epsilon_\alpha < 1/d \leq 1/|S|$, Theorem~\ref{thm:approx} with $T = k_n^{|S|}$ gives
$\boldsymbol{\psi}_S \in \Psi(k_n^{|S|}, B_S)$ with $S_t = S$ for all its basis functions such that
\begin{align*}
\Vert f_{0,S} - f_{\text{SCAN},S}(\cdot \mid \boldsymbol{\psi}_S, \epsilon_{\alpha}) \Vert_{\infty}
\leq \frac{4^{|S|} L |S|^{\rho/2}}{k_n^{\rho}} + 2^{|S|+1} F\, k_n^{|S|} e^{-k_n/2}
\leq \frac{C_S}{k_n^{\rho}},
\end{align*}
where $C_S$ depends only on $|S|$, $L$, $F$, and $\rho$, since $\sup_{k \geq 1} k^{|S| + \rho} e^{-k/2} < \infty$.
Moreover, by Theorem~\ref{thm:approx},
\begin{align*}
B_S = \max\Big\{ 2,\; \tfrac{1}{\mathbb{T}},\; 2^{|S|} F,\; 2p\, k_n,\; 40|S|\, k_n^{2},\; B_{\theta,S} \Big\} \leq C'_S\, k_n^{2},
\end{align*}
where $B_{\theta,S}$ is the last term in the definition of $B$ in Theorem~\ref{thm:approx} with $d$ replaced by $|S|$, and $C'_S$ depends only on $|S|$, $F$, $p$, $\mathbb{T}$, and $\epsilon_\alpha$.
Since $k_n^2 \leq 4 n^{2/(2\rho+d)}$ and $\kappa > 2/(2\rho+d)$, we have $B_S \leq B_n$ for all sufficiently large $n$.

Let $f_n^{*} := \sum_{S \in \mathcal{S}} f_{\text{SCAN},S}(\cdot \mid \boldsymbol{\psi}_S, \epsilon_{\alpha})$.
It uses $\sum_{S \in \mathcal{S}} k_n^{|S|} \leq |\mathcal{S}|\, k_n^{d} \leq 2^d |\mathcal{S}|\, n^{d/(2\rho+d)} \leq T_n$ basis functions.
We set $w_t = 0$, $\boldsymbol{\Theta}_t = \mathbf{0}$, $\beta_{j,t,1} = \beta_{j,t,2} = b_{j,t} = 0$, and $\gamma_{j,t} = 1$ for the remaining basis functions, which then do not contribute to the output.
Hence, $f_n^{*} \in \mathcal{F}(T_n, B_n, \epsilon_\alpha)$ for all sufficiently large $n$.
With $C_A := \sum_{S \in \mathcal{S}} C_S$,
\begin{align}
\Vert f_n^{*} - m \Vert_\infty
\leq \sum_{S \in \mathcal{S}} \Vert f_{\text{SCAN},S}(\cdot \mid \boldsymbol{\psi}_S, \epsilon_{\alpha}) - f_{0,S} \Vert_\infty
\leq \frac{C_A}{k_n^{\rho}}.
\label{eq:approx_sup}
\end{align}
Therefore, by (\ref{eq:approx_sup}),
\begin{align*}
\inf_{f \in \mathcal{F}_n} \Vert f - m \Vert_{2,\mu}^{2}
\leq \Vert f_n^{*} - m \Vert_{\infty}^{2}
\leq C_A^{2}\, k_n^{-2\rho}
\leq C_A^{2}\, n^{-\frac{2\rho}{2\rho + d}}.
\end{align*}
\\
$\newline$
\textbf{Step 2: Estimation term.}
Since $\mathcal{T}_{\beta_n}$ is 1-Lipschitz and the empirical $L_1$ norm is bounded by the sup norm,
we have, for any $\eta > 0$ and $\mathbf{x}_1^n$,
\begin{align*}
N_1(\eta, \mathcal{T}_{\beta_n}\mathcal{F}_n, \mathbf{x}_1^n) \leq N\big(\eta, \mathcal{F}(T_n, B_n,\epsilon_{\alpha}), \Vert \cdot \Vert_\infty\big).
\end{align*}
By Lemma~\ref{lem:covering} with $\eta = 1/(n\beta_n)$,
\begin{align*}
\log N_1\Big( \frac{1}{n\beta_n}, \mathcal{T}_{\beta_n}\mathcal{F}_n, \mathbf{x}_1^n \Big)
&\leq p T_n \log 2 + T_n (5p+2) \log\big( 1 + T_n (5p+2) B_n^{7}\, n \beta_n \big) \\
&\lesssim T_n \log n,
\end{align*}
since $T_n$, $B_n$, and $\beta_n$ grow at most polynomially in $n$.
Since $T_n \lesssim n^{d/(2\rho + d)}$, the estimation term is bounded by
\begin{align*}
\frac{c_3 (\log n)^{2} (C T_n \log n + 1)}{n}
\lesssim (\log n)^{3}\, \frac{n^{d/(2\rho+d)}}{n}
= (\log n)^{3}\, n^{-\frac{2\rho}{2\rho + d}}.
\end{align*}
\\
$\newline$
\textbf{Step 3: Conclusion.}
Combining Steps 1 and 2 with Lemma~\ref{lem:oracle} proves the claim for all $n \geq N_2$ for some $N_2$.
For $2 \leq n < N_2$, we have $\Vert \hat f_n - m \Vert_{2,\mu}^2 \leq (\beta_n + |\mathcal{S}| F)^2$, since $\Vert \hat f_n \Vert_\infty \leq \beta_n$.
Hence, the claim also holds for these finitely many $n$ after enlarging $C$.
\qed

\newpage

\subsection{Proof of Theorem~\ref{cor:selection}}
\label{app:identified}

The proof consists of four steps.
In Step 1, we show the consistency of the population identified components $\bar{\Pi}_S \hat{f}_n$ in the $L_2(\bar{\mu})$ and $L_2(\mu)$ norms.
In Step 2, we extend this result to the empirical $L_2$ norm.
In Step 3, we replace the population projection $\bar{\Pi}_S$ with the empirical projection $\Pi_S$.
In Step 4, we prove the selection consistency.
Throughout, $C_1, C_2, \ldots$ denote positive constants independent of $n$, and we assume $n \geq 2$ so that $r_n > 0$.

Let $\mu_j$ denote the marginal distribution of $X_j$ and let $\bar{\mu} := \mu_1 \otimes \cdots \otimes \mu_p$ be the product of the marginals.
In this section, we use the population counterpart $\bar{\Pi}_S$ of $\Pi_S$, defined by (\ref{eq:purify}) with $\nu$ replaced by $\bar{\mu}$.
We first compare the norms with respect to $\mu$ and $\bar{\mu}$.

Let $p_j$ denote the marginal density of $X_j$.
Since $p_j(x_j) = \int_{[0,1]^{p-1}} p_X(\mathbf{x})\, d\mathbf{x}_{-j}$ and $p_L \leq p_X \leq p_U$ on $[0,1]^p$, we have $p_L \leq p_j \leq p_U$ for each $j \in [p]$,
and hence $p_L^{\,p} \leq \prod_{j=1}^p p_j(x_j) \leq p_U^{\,p}$.
Since $\Vert g \Vert_{2,\mu}^2 = \int g(\mathbf{x})^2 p_X(\mathbf{x})\, d\mathbf{x}$ and $\Vert g \Vert_{2,\bar{\mu}}^2 = \int g(\mathbf{x})^2 \prod_{j=1}^p p_j(x_j)\, d\mathbf{x}$, it follows that
\begin{align}
c_L \Vert g \Vert_{2,\bar{\mu}}^2 \leq \Vert g \Vert_{2,\mu}^2 \leq c_U \Vert g \Vert_{2,\bar{\mu}}^2
\quad \text{for every measurable function } g,
\label{eq:norm_equiv}
\end{align}
where $c_L := p_L / p_U^{\,p}$ and $c_U := p_U / p_L^{\,p}$.
We also use the following two lemmas.

\begin{lemma}[Orthogonal decomposition, \cite{hooker_diag}]\label{lem:parseval}
For any $g \in L_2(\bar{\mu})$, the functions $\{\bar{\Pi}_S g\}_{S \subseteq [p]}$ are mutually orthogonal in $L_2(\bar{\mu})$ and $g = \sum_{S \subseteq [p]} \bar{\Pi}_S g$.
Hence,
\begin{align}
\Vert g \Vert_{2,\bar{\mu}}^2 = \sum_{S \subseteq [p]} \Vert \bar{\Pi}_S g \Vert_{2,\bar{\mu}}^2.
\label{eq:parseval}
\end{align}
\end{lemma}

\begin{lemma}[Theorem 19.3 of \cite{gyorfi2006distribution}]
\label{lem:gyorfi}
Let $\mathbf{X}, \mathbf{X}_1, \ldots, \mathbf{X}_n$ be i.i.d.\ random vectors with values in $\mathbb{R}^p$.
Let $K_1, K_2 \geq 1$ be constants and let $\mathcal{G}$ be a class of functions $g : \mathbb{R}^p \to \mathbb{R}$ such that
\begin{align}
|g(\mathbf{x})| \leq K_1, \qquad \mathbb{E}[g(\mathbf{X})^2] \leq K_2\, \mathbb{E}[g(\mathbf{X})].
\label{eq:gyo_1}
\end{align}
Let $0 < \eta < 1$ and $\zeta > 0$. Assume that
\begin{align}
\sqrt{n}\, \eta \sqrt{1-\eta} \sqrt{\zeta} \geq 288 \max\big\{2K_1, \sqrt{2K_2}\big\}
\label{eq:gyo_2}
\end{align}
and that, for all $\mathbf{x}_1, \ldots, \mathbf{x}_n \in \mathbb{R}^p$ and all $t \geq \zeta/8$,
\begin{align}
&\frac{\sqrt{n}\, \eta (1-\eta)\, t}{96\sqrt{2} \max\{K_1, 2K_2\}} \nonumber \\
&\geq \int_{\frac{\eta(1-\eta) t}{16 \max\{K_1, 2K_2\}}}^{\sqrt{t}}
\sqrt{\log N\Big(u, \Big\{ g \in \mathcal{G} : \tfrac{1}{n}\textstyle\sum_{i=1}^n g(\mathbf{x}_i)^2 \leq 16t \Big\}, \Vert \cdot \Vert_{1,n}\Big)}\, du.
\label{eq:gyo_3}
\end{align}
Then
\begin{align*}
\mathbb{P}\bigg( \sup_{g \in \mathcal{G}} \frac{\big| \mathbb{E}[g(\mathbf{X})] - \frac{1}{n}\sum_{i=1}^n g(\mathbf{X}_i) \big|}{\zeta + \mathbb{E}[g(\mathbf{X})]} > \eta \bigg)
\leq 60 \exp\bigg( - \frac{n \zeta \eta^2 (1-\eta)}{C_g \max\{K_1^2, K_2\}} \bigg)
\end{align*}
for some constant $C_g > 0$.
\end{lemma}

\begin{lemma}
\label{lem:emp_proj}
Under the conditions of Theorem~\ref{cor:selection}, there exists a constant $C_3 > 0$, independent of $n$, such that, for every $S \subseteq [p]$,
\begin{align}
\mathbb{E} \Vert \Pi_S \hat f_n - \bar{\Pi}_S \hat f_n \Vert_{2,n}^{2} \leq C_3\, r_n.
\label{eq:emp_proj}
\end{align}
\end{lemma}
\noindent\textbf{Proof.}
The lemma can be shown by the same argument as in the proof of Theorem~C.1 of \cite{park2025bayesian}, with the function class there replaced by $\mathcal{T}_{\beta_n}\mathcal{F}(T_n, B_n, \epsilon_\alpha)$.
In particular, the covering number of this class is bounded by Lemma~\ref{lem:covering}, and every function in the class is bounded by $\beta_n \lesssim \log n$ in the sup norm.
\qed
$\newline$
$\newline$
\textbf{Step 1: Consistency of the population identified components.}
We show that
\begin{align}
\mathbb{E} \sum_{S \subseteq [p]} \Vert \bar{\Pi}_S \hat f_n - f^{*}_{0,S} \Vert_{2,\bar{\mu}}^{2} &\leq \frac{C_1}{c_L}\, r_n,
\label{eq:step1} \\
\mathbb{E} \sum_{S \subseteq [p]} \Vert \bar{\Pi}_S \hat f_n - f^{*}_{0,S} \Vert_{2,\mu}^{2} &\leq \frac{c_U C_1}{c_L}\, r_n.
\label{eq:step1_mu}
\end{align}
By Theorem~\ref{thm:conv_rate}, $\mathbb{E} \Vert \hat f_n - f_0 \Vert_{2,\mu}^{2} \leq C_1 r_n$.
Hence, by the first inequality in (\ref{eq:norm_equiv}),
\begin{align*}
\mathbb{E} \Vert \hat f_n - f_0 \Vert_{2,\bar{\mu}}^{2} \leq \frac{1}{c_L}\, \mathbb{E} \Vert \hat f_n - f_0 \Vert_{2,\mu}^{2} \leq \frac{C_1}{c_L}\, r_n.
\end{align*}
Since $|\hat f_n| \leq \beta_n$ and $|f_0| \leq |\mathcal{S}| F$, the function $\hat f_n - f_0$ is bounded and belongs to $L_2(\bar{\mu})$.
Since $\bar{\Pi}_S$ is linear and $f^{*}_{0,S} = \bar{\Pi}_S f_0$, we have $\bar{\Pi}_S \hat f_n - f^{*}_{0,S} = \bar{\Pi}_S (\hat f_n - f_0)$, and Lemma~\ref{lem:parseval} gives
\begin{align*}
\sum_{S \subseteq [p]} \Vert \bar{\Pi}_S \hat f_n - f^{*}_{0,S} \Vert_{2,\bar{\mu}}^{2}
= \Vert \hat f_n - f_0 \Vert_{2,\bar{\mu}}^{2}.
\end{align*}
Taking expectations proves (\ref{eq:step1}).
Applying the second inequality in (\ref{eq:norm_equiv}) to each $\bar{\Pi}_S \hat f_n - f^{*}_{0,S}$ and summing over $S \subseteq [p]$, we obtain (\ref{eq:step1_mu}) from (\ref{eq:step1}).
Note that, unlike in $L_2(\bar{\mu})$, the functions $\{\bar{\Pi}_S g\}_{S \subseteq [p]}$ are not orthogonal in $L_2(\mu)$ in general, and hence only an inequality holds in (\ref{eq:step1_mu}).
\\
$\newline$
\textbf{Step 2: Extension to the empirical $L_2$ norm.}
We show that, for every $S \subseteq [p]$,
\begin{align}
\mathbb{E} \Vert \bar{\Pi}_S \hat f_n - f^{*}_{0,S} \Vert_{2,n}^{2} \leq C_2\, r_n,
\label{eq:step2}
\end{align}
where $\Vert \cdot \Vert_{2,n}$ is the empirical $L_2$ norm on the training inputs $\mathbf{X}_1, \ldots, \mathbf{X}_n$.
Fix $S \subseteq [p]$ and write $h_n := \bar{\Pi}_S \hat f_n - f^{*}_{0,S}$.
By (\ref{eq:step1_mu}), $\mathbb{E} \Vert h_n \Vert_{2,\mu}^{2} \leq (c_U C_1 / c_L)\, r_n$.
Since $h_n$ depends on the training inputs, $\mathbb{E} \Vert h_n \Vert_{2,n}^{2}$ cannot be compared with $\mathbb{E} \Vert h_n \Vert_{2,\mu}^{2}$ directly, and we use a uniform deviation inequality over a class containing $h_n$.

\medskip
\noindent\textit{(a) Uniform deviation.}
Let
\[
\mathcal{H}_n := \big\{ \bar{\Pi}_S \mathcal{T}_{\beta_n} f - f^{*}_{0,S} : f \in \mathcal{F}(T_n, B_n, \epsilon_\alpha) \big\},
\]
so that $h_n \in \mathcal{H}_n$.
Since each $\mathbb{E}_{\bar{\mu}}$ is a contraction in the sup norm and $\bar{\Pi}_S$ is a signed sum of $2^{|S|}$ such terms, $\Vert \bar{\Pi}_S g \Vert_\infty \leq 2^{|S|} \Vert g \Vert_\infty$ for any bounded $g$.
Since $\Vert \mathcal{T}_{\beta_n} f \Vert_\infty \leq \beta_n$ for every $f$ and $\Vert f_0 \Vert_\infty \leq |\mathcal{S}| F$, every $h \in \mathcal{H}_n$ satisfies $\Vert h \Vert_\infty \leq M_n := 2^{|S|}(\beta_n + |\mathcal{S}| F)$, and $1 \leq M_n \lesssim \log n$ for all sufficiently large $n$.
Consider the normalized class
\begin{align*}
\mathcal{G}_n := \big\{ h^2 / M_n^2 : h \in \mathcal{H}_n \big\}.
\end{align*}
Every $g \in \mathcal{G}_n$ satisfies $0 \leq g \leq 1$, so that $\mathbb{E}[g(\mathbf{X})^2] \leq \mathbb{E}[g(\mathbf{X})]$, and (\ref{eq:gyo_1}) holds with $K_1 = K_2 = 1$.

We next bound the covering number of $\mathcal{G}_n$.
Let $u > 0$ and $\delta := 2^{-|S|-1} u$, and let $\{f_1, \ldots, f_N\}$ be a $\delta$-cover of $\mathcal{F}(T_n, B_n, \epsilon_\alpha)$ in $\Vert \cdot \Vert_\infty$ with $N = N(\delta, \mathcal{F}(T_n, B_n, \epsilon_\alpha), \Vert \cdot \Vert_\infty)$.
For $f, f' \in \mathcal{F}(T_n, B_n, \epsilon_\alpha)$, let $h := \bar{\Pi}_S \mathcal{T}_{\beta_n} f - f^{*}_{0,S}$ and $h' := \bar{\Pi}_S \mathcal{T}_{\beta_n} f' - f^{*}_{0,S}$.
Since $|h|, |h'| \leq M_n$, $M_n \geq 1$, and $\mathcal{T}_{\beta_n}$ is 1-Lipschitz,
\begin{align*}
\frac{|h^2 - h'^2|}{M_n^2} \leq \frac{2|h - h'|}{M_n} \leq 2 \Vert \bar{\Pi}_S (\mathcal{T}_{\beta_n} f - \mathcal{T}_{\beta_n} f') \Vert_\infty \leq 2^{|S|+1} \Vert f - f' \Vert_\infty.
\end{align*}
Hence, $\{ (\bar{\Pi}_S \mathcal{T}_{\beta_n} f_j - f^{*}_{0,S})^2 / M_n^2 \}_{j=1}^N$ is a $u$-cover of $\mathcal{G}_n$ in $\Vert \cdot \Vert_\infty$, and thus in $\Vert \cdot \Vert_{1,n}$, since the empirical $L_1$ norm is bounded by the sup norm.
By Lemma~\ref{lem:covering}, for any $u > 0$ and $\mathbf{x}_1, \ldots, \mathbf{x}_n$,
\begin{align}
\log N\big(u, \mathcal{G}_n, \Vert \cdot \Vert_{1,n}\big)
&\leq \log N\big(2^{-|S|-1} u, \mathcal{F}(T_n, B_n, \epsilon_\alpha), \Vert \cdot \Vert_\infty\big) \nonumber \\
&\leq p T_n \log 2 + T_n(5p+2) \log\Big( 1 + \frac{2^{|S|+1} T_n (5p+2) B_n^7}{u} \Big).
\label{eq:emp_cover}
\end{align}

We apply Lemma~\ref{lem:gyorfi} with $K_1 = K_2 = 1$, $\eta = 1/2$, and $\zeta = \zeta_n := a_1 T_n \log n / n$ for a sufficiently large constant $a_1 > 0$.
\begin{itemize}
\item Condition (\ref{eq:gyo_2}) reads $\sqrt{n \zeta_n} \geq 288 \cdot 2^{5/2}$, which holds for all sufficiently large $n$ since $n \zeta_n = a_1 T_n \log n \to \infty$.
\item For (\ref{eq:gyo_3}), let $t \geq \zeta_n/8$.
The lower limit of the integral is $t/128 \geq \zeta_n/1024 \geq 1/n$ for all sufficiently large $n$.
Since $T_n$ and $B_n$ grow polynomially in $n$, (\ref{eq:emp_cover}) gives $\log N(u, \mathcal{G}_n, \Vert \cdot \Vert_{1,n}) \leq a_2 T_n \log n$ for all $u \geq 1/n$.
Hence, the right-hand side of (\ref{eq:gyo_3}) is at most $\sqrt{t}\, \sqrt{a_2 T_n \log n}$, while the left-hand side equals $\sqrt{n}\, t / (768\sqrt{2})$.
Thus, (\ref{eq:gyo_3}) holds whenever $t \geq a_3 T_n \log n / n$ with $a_3 := 2 \cdot 768^2 a_2$, which is guaranteed if $a_1 \geq 8 a_3$.
\end{itemize}
Let
\begin{align*}
A_n := \bigg\{ \sup_{g \in \mathcal{G}_n} \frac{\big| \mathbb{E}[g(\mathbf{X})] - \frac{1}{n}\sum_{i=1}^n g(\mathbf{X}_i) \big|}{\zeta_n + \mathbb{E}[g(\mathbf{X})]} \leq \frac{1}{2} \bigg\}.
\end{align*}
By Lemma~\ref{lem:gyorfi}, for all sufficiently large $n$,
\begin{align*}
\mathbb{P}(A_n^c) \leq 60 \exp\Big( - \frac{n \zeta_n}{8 C_g} \Big) = 60\, n^{-a_1 T_n/(8 C_g)} \leq \frac{1}{n}.
\end{align*}
On $A_n$, multiplying by $M_n^2$, every $h \in \mathcal{H}_n$ satisfies
\begin{align}
\Vert h \Vert_{2,n}^2 \leq \frac{3}{2} \Vert h \Vert_{2,\mu}^2 + \frac{M_n^2 \zeta_n}{2}.
\label{eq:step2b}
\end{align}
Since the supremum is taken over the whole class $\mathcal{H}_n$, (\ref{eq:step2b}) applies to the data-dependent function $h_n$.

\medskip
\noindent\textit{(b) Expectation.}
Using (\ref{eq:step2b}) on $A_n$ and $\Vert h_n \Vert_{2,n}^2 \leq M_n^2$ on $A_n^c$,
\begin{align*}
\mathbb{E} \Vert h_n \Vert_{2,n}^2
\leq \frac{3}{2}\, \mathbb{E} \Vert h_n \Vert_{2,\mu}^2 + \frac{M_n^2 \zeta_n}{2} + M_n^2\, \mathbb{P}(A_n^c).
\end{align*}
The first term is bounded by (\ref{eq:step1_mu}).
Since $M_n^2 \lesssim (\log n)^2$ and $T_n \lesssim n^{d/(2\rho+d)}$,
\begin{align*}
M_n^2 \zeta_n \lesssim (\log n)^3\, \frac{T_n}{n} \lesssim r_n,
\qquad
M_n^2\, \mathbb{P}(A_n^c) \lesssim \frac{(\log n)^2}{n} \lesssim r_n.
\end{align*}
This proves (\ref{eq:step2}) for all $n \geq N_0$ for some $N_0$.
For $2 \leq n < N_0$, (\ref{eq:step2}) follows by enlarging $C_2$, since $\mathbb{E} \Vert h_n \Vert_{2,n}^2 \leq M_n^2$ and $r_n > 0$.
\\
$\newline$
\textbf{Step 3: From the population projection to the empirical projection.}
For nonempty $S \subseteq [p]$, let
\begin{align*}
\Delta_S := \Vert \Pi_S \hat f_n - f^{*}_{0,S} \Vert_{2,n}.
\end{align*}
By the triangle inequality,
\begin{align*}
\Delta_S \leq \Vert \Pi_S \hat f_n - \bar{\Pi}_S \hat f_n \Vert_{2,n} + \Vert \bar{\Pi}_S \hat f_n - f^{*}_{0,S} \Vert_{2,n}.
\end{align*}
Hence, by $(a+b)^2 \leq 2a^2 + 2b^2$, (\ref{eq:emp_proj}), and (\ref{eq:step2}),
\begin{align}
\mathbb{E} \Delta_S^{2} \leq 2(C_2 + C_3)\, r_n =: C_4\, r_n.
\label{eq:step3}
\end{align}
\\
$\newline$
\textbf{Step 4: Selection consistency.}
Let
\begin{align*}
m^{*} := \min_{S \in \mathcal{S}^{*}} \Vert f^{*}_{0,S} \Vert_{2,\mu}.
\end{align*}
By the first inequality in (\ref{eq:norm_equiv}), $\Vert f^{*}_{0,S} \Vert_{2,\mu}^2 \geq c_L \Vert f^{*}_{0,S} \Vert_{2,\bar{\mu}}^2 > 0$ for each $S \in \mathcal{S}^{*}$.
Since $\mathcal{S}^{*}$ is finite, we have $m^{*} > 0$.
Since each $f^{*}_{0,S}$ is a fixed bounded function, the law of large numbers implies that the event
\begin{align*}
D_n := \big\{ \Vert f^{*}_{0,S} \Vert_{2,n} \geq m^{*}/2 \ \text{for all } S \in \mathcal{S}^{*} \big\}
\end{align*}
satisfies $\mathbb{P}(D_n) \to 1$.
Since $\tau_{f,n} \to 0$, we have $\tau_{f,n} \leq m^{*}/4$ for all sufficiently large $n$.
For such $n$, on the event $D_n \cap \Omega_n$ with $\Omega_n := \{ \max_{S \neq \emptyset} \Delta_S < \tau_{f,n} \}$, the following hold.
\begin{itemize}
\item If $S \in \mathcal{S}^{*}$, then by the reverse triangle inequality,
$\Vert \Pi_S \hat f_n \Vert_{2,n} \geq \Vert f^{*}_{0,S} \Vert_{2,n} - \Delta_S > m^{*}/2 - \tau_{f,n} \geq \tau_{f,n}$,
and hence $S \in \hat{\mathcal{S}}_n$.
\item If $S \notin \mathcal{S}^{*}$ and $S \neq \emptyset$, then $f^{*}_{0,S} = 0$, so that $\Vert \Pi_S \hat f_n \Vert_{2,n} = \Delta_S < \tau_{f,n}$, and hence $S \notin \hat{\mathcal{S}}_n$.
\end{itemize}
Therefore, $\{ \hat{\mathcal{S}}_n \neq \mathcal{S}^{*} \} \subseteq D_n^c \cup \Omega_n^c$ and $\Omega_n^c \subseteq \{ \sum_{S \neq \emptyset} \Delta_S^2 \geq \tau_{f,n}^2 \}$.
By Markov's inequality and (\ref{eq:step3}) summed over the $2^p - 1$ nonempty subsets,
\begin{align*}
\mathbb{P}\big( \hat{\mathcal{S}}_n \neq \mathcal{S}^{*} \big)
&\leq \mathbb{P}(D_n^c) + \frac{\mathbb{E} \sum_{S \neq \emptyset} \Delta_S^{2}}{\tau_{f,n}^{2}} \\
&\leq \mathbb{P}(D_n^c) + \frac{2^p C_4\, r_n}{\tau_{f,n}^{2}} \to 0,
\end{align*}
where the convergence follows from $\mathbb{P}(D_n) \to 1$ and $\tau_{f,n}^2 / r_n \to \infty$.
\qed

\subsection{Screening property of the discovered components}
\label{app:screening}

\begin{proposition}[Screening property of the discovered components]\label{prop:screening}
Suppose that the conditions of Theorem~\ref{cor:selection} hold.
Let $\hat{\boldsymbol{\psi}}_n \in \Psi(T_n, B_n)$ be any measurable choice of parameter with
$\hat f^{\mathrm{raw}}_n = f_{\text{SCAN}}(\cdot \mid \hat{\boldsymbol{\psi}}_n, \epsilon_\alpha)$,
let $\hat{\mathcal{A}}_n := \{ S_t(\hat{\boldsymbol{\psi}}_n) : t \in [T_n],\ \hat w_t \neq 0 \}$,
and let $E_n := \{ \Vert \hat f^{\mathrm{raw}}_n \Vert_\infty \leq \beta_n \}$ be the event that the clipping is inactive.
Then
\begin{align*}
\mathbb{P}\big( \{\mathcal{S}^{*} \not\subseteq \mathcal{D}(\hat{\mathcal{A}}_n)\} \cap E_n \big) \to 0 .
\end{align*}
In particular, if $\mathbb{P}(E_n) \to 1$, then $\mathbb{P}( \mathcal{S}^{*} \subseteq \mathcal{D}(\hat{\mathcal{A}}_n) ) \to 1$.
\end{proposition}
$\newline$
\textbf{Proof.}
Since $\hat f^{\mathrm{raw}}_n = \sum_{t : \hat w_t \neq 0} \hat w_t \mathcal{M}_t$ and each $\mathcal{M}_t$ depends on $\mathbf{x}$ only through $\mathbf{x}_{S_t(\hat{\boldsymbol{\psi}}_n)}$, Proposition~\ref{prop:projection-0} applied with $\bar{\mu}$ in place of $\nu$ gives
\begin{align}
\bar{\Pi}_S \hat f^{\mathrm{raw}}_n = 0 \qquad \text{for every } S \notin \mathcal{D}(\hat{\mathcal{A}}_n).
\label{eq:screen_zero}
\end{align}
On $E_n$, we have $\hat f_n = \mathcal{T}_{\beta_n}\hat f^{\mathrm{raw}}_n = \hat f^{\mathrm{raw}}_n$, and hence $\bar{\Pi}_S \hat f_n = 0$ for every $S \notin \mathcal{D}(\hat{\mathcal{A}}_n)$.

Let $\bar{m}^{*} := \min_{S \in \mathcal{S}^{*}} \Vert f^{*}_{0,S} \Vert_{2,\bar{\mu}}$.
Since $\mathcal{S}^{*}$ is finite and $\Vert f^{*}_{0,S} \Vert_{2,\bar{\mu}} > 0$ for each $S \in \mathcal{S}^{*}$, we have $\bar{m}^{*} > 0$.
On $\{\mathcal{S}^{*} \not\subseteq \mathcal{D}(\hat{\mathcal{A}}_n)\} \cap E_n$, take $S \in \mathcal{S}^{*} \setminus \mathcal{D}(\hat{\mathcal{A}}_n)$. Then
\begin{align*}
\Vert \bar{\Pi}_S \hat f_n - f^{*}_{0,S} \Vert_{2,\bar{\mu}} = \Vert f^{*}_{0,S} \Vert_{2,\bar{\mu}} \geq \bar{m}^{*}.
\end{align*}
Hence,
\begin{align*}
\{ \mathcal{S}^{*} \not\subseteq \mathcal{D}(\hat{\mathcal{A}}_n) \} \cap E_n
\subseteq
\Big\{ \sum_{S \subseteq [p]} \Vert \bar{\Pi}_S \hat f_n - f^{*}_{0,S} \Vert_{2,\bar{\mu}}^{2} \geq (\bar{m}^{*})^{2} \Big\}.
\end{align*}
By Markov's inequality and (\ref{eq:step1}),
\begin{align*}
\mathbb{P}\big( \{\mathcal{S}^{*} \not\subseteq \mathcal{D}(\hat{\mathcal{A}}_n)\} \cap E_n \big)
\leq \frac{C_1}{c_L\, (\bar{m}^{*})^{2}}\, r_n \to 0.
\end{align*}
The second claim follows from
$\mathbb{P}( \mathcal{S}^{*} \not\subseteq \mathcal{D}(\hat{\mathcal{A}}_n) ) \leq \mathbb{P}( \{\mathcal{S}^{*} \not\subseteq \mathcal{D}(\hat{\mathcal{A}}_n)\} \cap E_n ) + \mathbb{P}(E_n^c)$.
\qed

\newpage

\section{Experimental Details}

The gradient-based neural models are trained using the Adam optimizer, while XGB uses boosting.
The benchmark experiments use RTX 3090 or RTX 4090 GPUs with 24 GB of memory.

\subsection{Datasets}
\label{app:data}

\begin{table}[h]
\large
\centering
\caption{\textbf{Summary of benchmark datasets.}}
\label{Table : Dataset}
\begin{tabular}{lrrcc}
\toprule
Dataset & Size & \# Features & Problem & \# Classes \\
\midrule
\midrule
\textsc{Calhousing} (\cite{calhousing}) & 20,640  & 8   & Regression     & - \\
\textsc{Wine} (\cite{wine_quality})     & 4,898   & 11  & Regression     & - \\
\textsc{Microsoft} (\cite{micro})       & 964,933 & 136 & Regression     & - \\
\textsc{Yahoo} (\cite{yahoo})           & 709,877 & 699 & Regression     & - \\
\midrule
\textsc{Churn} (\cite{churn})           & 7,043   & 19  & Classification & 2 \\
\textsc{Credit} (\cite{credit})         & 284,807 & 30  & Classification & 2 \\
\textsc{Letter} (\cite{Letter})         & 20,000  & 16  & Classification & 2 \\
\textsc{Drybean} (\cite{drybean})       & 13,611  & 16  & Classification & 7 \\
\textsc{Madelon} (\cite{madelon_171})   & 2,600   & 500 & Classification & 2 \\
\midrule
Breast cancer gene (\cite{hatzis2011genomic}) & 488 & 12,502 & Classification & 2 \\
\bottomrule
\end{tabular}%
\end{table}
Table~\ref{Table : Dataset} summarizes the datasets used in the experiments, and each dataset is described below. \\
$\newline$
\textbf{\textsc{Calhousing}  (\cite{scikit-learn}).} \textsc{Calhousing} dataset contains information about housing prices in California, based on data from the 1990 U.S. Census.
It includes features such as the median income, house age, number of rooms, number of bedrooms, population, and geographic location (latitude and longitude) of various districts. The target variable is the median house value for each district.
We use \textsc{Calhousing} dataset in scikit-learn python package \cite{scikit-learn}.
\\
$\newline$
\textbf{\textsc{Wine}.}
\textsc{Wine} dataset consists of white Vinho Verde wine samples from northern Portugal.
The objective is to predict wine quality based on results from physicochemical tests.
We use \textsc{Wine} dataset in \\
\href{https://archive.ics.uci.edu/dataset/186/wine+quality}{https://archive.ics.uci.edu/dataset/186/wine+quality}.
\\
$\newline$
\textbf{\textsc{Microsoft}.}
\textsc{Microsoft} dataset consists of features extracted from query-URL pairs.
The target relevance score ranges from 0 (irrelevant) to 4 (very relevant).
We use \textsc{Microsoft} dataset in \href{https://www.microsoft.com/en-us/research/project/mslr/}{https://www.microsoft.com/en-us/research/project/mslr/}.
\\
$\newline$
\textbf{\textsc{Yahoo}.}
Like \textsc{Microsoft}, \textsc{Yahoo} dataset contains features extracted from query-URL pairs.
The target relevance score ranges from 0 to 4.
We use \textsc{Yahoo} dataset in \\
\href{https://webscope.sandbox.yahoo.com/catalog.php?datatype=c&guccounter=1}{https://webscope.sandbox.yahoo.com/catalog.php?datatype=c\&guccounter=1}
\\
$\newline$
\textbf{\textsc{Churn}.}
\textsc{Churn} task predicts whether a customer cancels a telecommunications subscription.
We use \textsc{Churn} dataset in \href{https://www.kaggle.com/datasets/blastchar/telco-customer-churn}{https://www.kaggle.com/datasets/blastchar/telco-customer-churn}
\\
$\newline$
\textbf{\textsc{Credit}.}
\textsc{Credit} task predicts whether a credit-card transaction is fraudulent.
Most input features are principal component scores released to protect sensitive transaction information.
We use \textsc{Credit} dataset in \href{https://www.kaggle.com/datasets/mlg-ulb/creditcardfraud}{https://www.kaggle.com/datasets/mlg-ulb/creditcardfraud}.
\\
$\newline$
\textbf{\textsc{Letter}.}
The goal is to classify a large set of black-and-white rectangular pixel images as one of the 26 uppercase letters of the English alphabet. The character images were generated using 20 different fonts, and each letter in these fonts was randomly distorted, resulting in a dataset of 20,000 unique samples.
To convert this dataset into a binary classification problem, the target variable is changed to 0 if it falls between A and M, and 1 otherwise.
We use \textsc{Letter} dataset in \href{https://archive.ics.uci.edu/dataset/59/letter+recognition}{https://archive.ics.uci.edu/dataset/59/letter+recognition}.
\\
$\newline$
\textbf{\textsc{Drybean}.}
\textsc{Drybean} task classifies beans into seven varieties using 16 measurements of size and shape extracted from images.
We use \textsc{Drybean} dataset in \\
\href{https://archive.ics.uci.edu/dataset/602/dry+bean+dataset}{https://archive.ics.uci.edu/dataset/602/dry+bean+dataset}.
\\
$\newline$
\textbf{\textsc{Madelon}.}
\textsc{Madelon} is an artificial binary classification dataset from the NIPS 2003 feature selection challenge (\cite{madelon_171}).
Among its 500 features, only 20 are relevant, and the class label depends on their interactions rather than their main effects.
We use \textsc{Madelon} dataset from the
\href{https://archive.ics.uci.edu/dataset/171/madelon}{UCI Machine Learning Repository}.

\subsection{Implementation of Baseline Models}
$\newline$
\textbf{Neural Additive Models (NAM, \cite{agarwal2021neural}).}
For NAM with main effects, we use the official implementation at \\
\href{https://github.com/AmrMKayid/nam}{https://github.com/AmrMKayid/nam}.
Additionally, we extend the code to implement NA$^{2}$M, which estimates up to second-order components.
\\
$\newline$
\textbf{Neural Basis Models (NBM, \cite{radenovic2022neural}).}
For NBM, we use the official implementation at
\href{https://github.com/facebookresearch/nbm-spam}{https://github.com/facebookresearch/nbm-spam}.
\\
$\newline$
\textbf{ANOVA-TPNN (\cite{park2025tensor}).}
For ANOVA-TPNN, we use the official implementation at \href{https://github.com/ParkSeokhun/ANOVA-TPNN}{https://github.com/ParkSeokhun/ANOVA-TPNN}.
\\
$\newline$
\textbf{NODE-GAM (\cite{chang2021node}).}
For NODE-GAM, we use the official implementation at \href{https://github.com/zzzace2000/nodegam}{https://github.com/zzzace2000/nodegam}.
\\
$\newline$
\textbf{XGB (\cite{chen2016xgboost}).}
We use the XGB Python package (\cite{chen2016xgboost}).
\\
$\newline$
\textbf{Multilayer Perceptron (MLP, \cite{hornik1989multilayer}).}
We implement MLP in PyTorch (\cite{NEURIPS2019_9015}).
\\
$\newline$
\textbf{Kolmogorov-Arnold Network (KAN, \cite{liu2024kan}).}
We use the official pykan package for KAN (\url{https://github.com/KindXiaoming/pykan}).
\\
$\newline$
\textbf{Explainable Boosting Machine (EBM, \cite{nori2019interpretml}).}
We use the official InterpretML package for EBM (\url{https://github.com/interpretml/interpret}).


\subsection{Hyperparameters}
\label{app:hyper}

In all experiments on \textsc{Calhousing}, \textsc{Wine}, \textsc{Churn}, \textsc{Credit}, \textsc{Letter}, and \textsc{Drybean}, we set the batch size to 4096, and for \textsc{Microsoft} and \textsc{Yahoo}, we use the maximal batch size that fits in our computational environment.
Unless stated otherwise, gradient-based neural models use Adam with a learning rate of $5 \times 10^{-3}$.

For NAM and NBM, the maximum component order is selected from $\{1,2\}$.
In NAM, the hidden layers of each component network have dimensions [64, 32, 16] for \textsc{Microsoft} and \textsc{Yahoo} datasets and [64, 64, 32] for the other datasets when only the main effects are estimated.
When the components up to the second order are estimated, we use [64, 32, 16] for \textsc{Credit} and \textsc{Drybean} datasets, [64, 16, 8] for \textsc{Microsoft} and \textsc{Yahoo} datasets, and [64, 64, 32] for the other datasets.
In NBM, the hidden layers of the basis neural network have dimensions [256, 256, 128], which are identical to those used in \cite{radenovic2022neural}.

For ANOVA-TPNN, we use the hyperparameter settings reported in the original papers (\cite{park2025tensor}).

For SCAN, NODE-GAM, XGB, KAN, and MLP, hyperparameters are selected by grid search using the average validation performance over ten random splits.
Models are fitted on the training subset of each split, and the test subsets are used only for the reported prediction evaluation.
\\
$\newline$
\textbf{Range of hyperparameters for NODE-GAM.}
For \textsc{Calhousing}, \textsc{Wine}, \textsc{Churn}, \textsc{Letter}, \textsc{Credit}, \textsc{Drybean} and \textsc{Madelon} datasets, we consider the range of hyperparameters for grid search as

\begin{itemize}
    \item Number of layers: \{2, 4, 8\}
    \item Tree depth: \{6, 8\}
    \item Number of trees per layer: \{256, 512\}
\end{itemize}
For \textsc{Microsoft} and \textsc{Yahoo} datasets, we conduct the grid search within the following ranges due to limitations in our computational resources.

\begin{itemize}
    \item Number of layers: \{2, 4\}
    \item Tree depth: \{6\}
    \item Number of trees per layer: \{256, 512\}
\end{itemize}
$\newline$
\textbf{Range of hyperparameters for SCAN.}
In the benchmark experiments, the temperatures are annealed with $\mathbb{T}_0 = \mathbb{T}_{g,0} = 3$ and $E_a$ set to 60\% of the training epochs (see Section~\ref{app:annealing}),
the truncation threshold is set to $\epsilon_\alpha = 0.01$,
and the post-hoc identification uses the 100 empirical quantiles of each feature.
For \textsc{Calhousing}, \textsc{Wine}, \textsc{Churn}, \textsc{Letter}, \textsc{Credit}, \textsc{Drybean} and $\textsc{Madelon}$,
we select the hyperparameters by grid search on the validation data over
\begin{itemize}
    \item the final temperature $\mathbb{T}_{\text{end}}$: \{0.03, 0.1, 0.3\},
    \item the final global temperature $\mathbb{T}_{g,\text{end}}$: \{0.01, 0.02, 0.03\},
    \item the number of basis functions $T$: \{100, 200, 500, 1000\},
\end{itemize}
where only the pairs with $\mathbb{T}_{g,\text{end}} < \mathbb{T}_{\text{end}}$ are considered.
For \textsc{Microsoft} and \textsc{Yahoo}, due to limited computational resources, we use the same grids for $\mathbb{T}_{\text{end}}$ and $\mathbb{T}_{g,\text{end}}$ and $T \in \{100, 300, 500\}$.
For \textsc{Madelon}, we use $T = 200$, a learning rate of $10^{-2}$, and a weight decay of $10^{-3}$, and select $(\mathbb{T}_{\text{end}}, \mathbb{T}_{g,\text{end}})$ over the grid.\\
$\newline$
\textbf{Range of hyperparameters for XGB.}
For \textsc{Calhousing}, \textsc{Wine}, \textsc{Churn}, \textsc{Letter}, \textsc{Credit}, \textsc{Drybean} and \textsc{Madelon} datasets, we consider the range of hyperparameters for grid search as

\begin{itemize}
    \item Number of trees: \{50,100,200,300,400,500,600,700,800,900,1000\}
    \item Maximum depth: \{3 , 5 , 7\}
    \item Learning rate: \{0.0001, 0.005, 0.01, 0.05 , 0.1\}
\end{itemize}
For \textsc{Microsoft} and \textsc{Yahoo} datasets, we conduct the grid search within the following ranges due to limitations in our computational resources.

\begin{itemize}
    \item Number of trees: \{100,200,300,400,500\}
    \item Maximum depth: \{7\}
    \item Learning rate: \{0.001, 0.01, 0.1\}
\end{itemize}
$\newline$
\textbf{Range of hyperparameters for MLP.}
For all datasets, we select the architecture that has optimal prediction performance among the following:
\begin{itemize}
    \item two hidden layers with 2048 and 1024 units
    \item three hidden layers with 1024,512 and 512 units
    \item five hidden layers with 128, 128, 64, 64, and 64 units
\end{itemize}
$\newline$
\textbf{Range of hyperparameters for KAN.}
For all datasets, we select the architecture that has optimal prediction performance among the following:
\begin{itemize}
    \item one hidden layer with 50 units
    \item one hidden layer with 100 units
    \item two hidden layers with 100 and 50 units
    \item two hidden layers with 100 and 100 units
\end{itemize}
$\newline$
\textbf{Range of hyperparameters for EBM.}
For all datasets, we select the hyperparameters that has optimal prediction performance among the following:
\begin{itemize}
    \item learning rate: \{0.005, 0.01, 0.015, 0.03\}
    \item maximum number of leaves per tree: \{2, 3, 4\}
    \item minimum number of samples per leaf: \{2, 4, 10\}
    \item number of pairwise interactions: \{0, $p$, $3p$, $5p$\}, where $p$ is the number of input features
\end{itemize}
The remaining hyperparameters are fixed to the default values of the \texttt{interpret} package (version 0.7.8).

\subsection{Experimental details for the Breast Cancer Application}
\label{app:application}
$\newline$
\textbf{Data and preprocessing.}
We use the published processed microarray values and the GPL96 platform annotation.
Probes without an unambiguous single gene symbol are excluded, and probes mapped to the same gene are summarized by their median within each patient.
Only patients with an observed \texttt{pCR} or \texttt{RD} label in \texttt{pathologic\_response\_pcr\_rd} are included.
Clinical variables, survival outcomes, and previously supplied response predictions are excluded from the inputs.
With the selected hyperparameters, each model is trained on the development cohort with five different random seeds and evaluated separately on the external cohort.
\\
$\newline$
\textbf{Hyperparameters.}
For SCAN, NODE-GAM and EBM, hyperparameters are selected by three-fold cross-validation over the candidate values listed in Section~\ref{app:hyper}. 
\\
$\newline$
\textbf{Component identification.}
For visualization, we retain only the identified components whose importance exceeds $0.003$, which changes the predictions by less than $0.03$ on the log-odds scale and leaves 61 components on 53 genes.

\subsection{Temperature Annealing}
\label{app:annealing}
$\newline$
\textbf{Motivation.}
If the training starts with small temperatures, the softmax in (\ref{eq:dual_temp}) becomes nearly deterministic from the beginning.
In that case, the features selected by each basis function are largely determined by the random initialization, and they are hard to correct afterwards.
To avoid this, we start with large temperatures, which allow each basis function to explore many features, and gradually decrease them so that the selection concentrates on a few relevant features.
\\
$\newline$
\textbf{Temperature annealing.}
We start with a large temperature, so that every basis function
initially uses almost all features ($\alpha_{\cdot,t}$ is nearly uniform), and decrease it exponentially
during the first $E_a$ epochs. It is then kept fixed:
\begin{align*}
\mathbb{T}(k) = \mathbb{T}_{0} \left( \frac{\mathbb{T}_{\text{end}}}{\mathbb{T}_{0}} \right)^{\min(k/E_a,\, 1)},
\end{align*}
where $k$ is the training epoch.
When the global component-selection parameter $\mathbf{g}$ is used, the global temperature $\mathbb{T}_g$ is decreased in the
same way, from $\mathbb{T}_{g,0}$ to $\mathbb{T}_{g,\text{end}}$.
Unless stated otherwise, we set $\mathbb{T}_0 = \mathbb{T}_{g,0} = 3$ and $E_a$ to about 60\% of the
training epochs (e.g., $E_a = 500$ out of 800 epochs for \textsc{Madelon} dataset).
The final values $(\mathbb{T}_{\text{end}}, \mathbb{T}_{g,\text{end}})$ are chosen by grid search on the
validation data.
We also apply weight decay to all parameters.
\\
$\newline$
\textbf{Choice of the final temperatures.}
The two final temperatures $\mathbb{T}_{\text{end}}$ and $\mathbb{T}_{g,\text{end}}$ control local component-selection and global component-selection, respectively.
The following choices describe the benchmark experiments. 
In the benchmark grid, we consider pairs with $\mathbb{T}_{g,\text{end}} < \mathbb{T}_{\text{end}}$.
This is because the selection of which features are used by the model as a whole should be sharper than the selection of features within each basis function: irrelevant features should be excluded from the entire model, whereas each basis function should retain enough flexibility to combine the relevant features into interactions.
$\mathbb{T}_{\text{end}}$ controls the trade-off between interpretability and prediction accuracy.
A smaller $\mathbb{T}_{\text{end}}$ yields lower-order components that are easier to interpret, whereas a larger $\mathbb{T}_{\text{end}}$ allows higher-order components and typically improves the prediction performance.
In practice, we select $(\mathbb{T}_{\text{end}}, \mathbb{T}_{g,\text{end}})$ by grid search on the validation data over
$\mathbb{T}_{\text{end}} \in \{0.03, 0.1, 0.3\}$ and $\mathbb{T}_{g,\text{end}} \in \{0.01, 0.02, 0.03\}$.
Sensitivity to these choices is examined in Section~\ref{app:sen_final}.

\subsection{Synthetic Datasets}
\label{app:syn_data}

\begin{table}[h]
\centering
\scriptsize
\caption{\textbf{Closed-form expressions of the synthetic functions $f^{(1)}$, $f^{(2)}$, and $f^{(3)}$.}}
\label{table:def_syn}
\begin{tabular}{l c}
\toprule
Target & Closed-form expression \\
\midrule
$f^{(1)}(\mathbf{x})$ &
$\pi^{x_1x_2}\sqrt{2|x_3|}-\arcsin(0.5x_4)+\log(|x_3+x_5|+1)
+\dfrac{x_9}{1+|x_{10}|}\sqrt{\dfrac{x_7}{1+|x_8|}}-x_2x_7$ \\[1.2ex]
$f^{(2)}(\mathbf{x})$ &
$x_1x_2+2^{x_3+x_5+x_6}+2^{x_3+x_4+x_5+x_7}
+\sin\!\big(x_7\sin(x_8+x_9)\big)+\arccos(0.9x_{10})$ \\[1.2ex]
$f^{(3)}(\mathbf{x})$ &
$\tanh(x_1x_2+x_3x_4)\sqrt{|x_5|}+\exp(x_5+x_6)
+\log\!\big((x_6x_7x_8)^2+1\big)+x_9x_{10}+\dfrac{1}{1+|x_{10}|}$ \\
\bottomrule
\end{tabular}
\end{table}

\begin{table}[h]
\centering
\scriptsize
\caption{\textbf{Sampling distributions of the input features used to generate each synthetic function.}}
\label{table:dit_syn}
\begin{tabular}{l c}
\toprule
Target & Input distribution \\
\midrule
$f^{(1)}(\mathbf{x})$ &
\begin{tabular}[c]{@{}c@{}}
$X_j \overset{\text{iid}}{\sim} \mathrm{U}(0,1)$ for $j\in\{1,2,3,6,7,9\}$,
$X_j \overset{\text{iid}}{\sim} \mathrm{U}(0.6,1)$ for $j\in\{4,5,8,10\}$ \\
$X_j \overset{\text{iid}}{\sim} \mathrm{U}(-1,1)$ for $j=11,\ldots,50$
\end{tabular} \\[2.2ex]
$f^{(2)}(\mathbf{x})$ & $X_1,\ldots,X_{50} \overset{\text{iid}}{\sim} \mathrm{U}(-1,1)$ \\
$f^{(3)}(\mathbf{x})$ & $X_1,\ldots,X_{50} \overset{\text{iid}}{\sim} \mathrm{U}(-1,1)$ \\
\bottomrule
\end{tabular}
\end{table}

$\newline$
\textbf{Synthetic Data Generation.}
We generate synthetic datasets from the regression model $Y = f^{(k)}(\mathbf{x}) + \epsilon$, where $\epsilon \sim N(0,\sigma^{2}_{\epsilon})$ and $\mathbf{x} \in \mathbb{R}^{50}$.
The true prediction models $f^{(k)}$ for $k=1,2,3$ are adopted from \cite{park2025bayesian,tsang2017detecting} and specified in Table~\ref{table:def_syn}.
Input features are sampled according to the distributions given in Table~\ref{table:dit_syn}, where variables 1-10 are informative (influencing the output) and variables 11-50 are non-informative noise features.
The noise variance $\sigma^{2}_{\epsilon}$ is set such that the signal-to-noise ratio equals 5.\\
$\newline$
\textbf{Evaluation Methodology.}
Following \cite{park2025tensor}, we evaluate component selection by AUROC.
For each true model $f^{(k)}$, $k \in \{1,2,3\}$, we compute AUROC from the pairs $(\Vert \hat f^{(k)}_S \Vert_{2,n}, r^{(k)}_S)$ over all $S \subseteq [p]$ with $|S| = k$.
Here, $\hat f^{(k)}_S$ is the estimated component for $S$, and $r^{(k)}_S = \mathbb{I}(\Vert \Pi_S f^{(k)} \Vert_{2,n} > 0)$ indicates whether the true component is active.
For SCAN, we use the identified component $\hat f^{(k)}_S = \Pi_S \hat f^{(k)}$.

\newpage

\section{Ablation Studies}

We examine four design choices: the number of basis functions $T$, the initial temperature $\mathbb{T}_0$, the final temperatures, and the univariate unit function.
The experiments use \textsc{Calhousing} or \textsc{Madelon}, with test RMSE or AUROC averaged over ten random splits, except for Section~\ref{app:sen_final}, which reports the results of a single split.

\subsection{Number of Basis Functions $T$}
To measure the orders of the estimated components, we define the effective order as the average order of the basis functions, i.e., $\frac{1}{T}\sum_{t=1}^{T} |S_t|$.
Table~\ref{tab:abl_T} shows the results as $T$ varies.
The RMSE decreases as $T$ increases, with smaller gains beyond $T=100$.
The effective order also decreases as $T$ increases.
This pattern is consistent with a larger basis expansion distributing the fitted function across more lower-order terms.
In this experiment, increasing $T$ therefore improves prediction while reducing the average component index set order.

\begin{table}[h]
\centering
\caption{\footnotesize \textbf{Effect of the number of basis functions $T$.}}
\label{tab:abl_T}
\begin{tabular}{ccccccc}
\toprule
$T$ & 10 & 25 & 50 & 100 & 200 & 400 \\
\midrule
RMSE $\downarrow$ & 0.546 & 0.523 & 0.509 & 0.503 & 0.501 & 0.498 \\
Standard deviation & (0.018) & (0.016) &   (0.014) & (0.016) & (0.016) & (0.018) \\
Effective order & 2.3 & 2.0 & 1.7 & 1.5 & 1.3 & 1.3 \\
\bottomrule
\end{tabular}
\end{table}

\subsection{Initial Temperature $\mathbb{T}_0$}
Table~\ref{tab:abl_T0} shows the results as the initial temperature $\mathbb{T}_0$ varies, where the temperature is annealed to the final value $\mathbb{T}_{\text{end}} = 0.1$.
Note that $\mathbb{T}_0 = 0.1$ corresponds to training without annealing.
The RMSE improves when $\mathbb{T}_0 \geq 3$ and remains nearly unchanged for larger values.
With a large $\mathbb{T}_0$, the component-selection weights $\alpha_{j,t}$ are nearly uniform at the beginning of training, so that each basis function can explore all features before the selection becomes sparse.

\begin{table}[h]
\centering
\caption{\footnotesize \textbf{Effect of the initial temperature $\mathbb{T}_0$ of the annealing schedule.}}
\label{tab:abl_T0}
\begin{tabular}{ccccccc}
\toprule
$\mathbb{T}_0$ & 0.1 & 0.3 & 1 & 3 & 10 & 30 \\
\midrule
RMSE $\downarrow$ & 0.515 & 0.516 & 0.512 & 0.501 & 0.503 & 0.499 \\
\bottomrule
\end{tabular}
\end{table}

\subsection{Sensitivity to the Final Temperatures}
\label{app:sen_final}
\begin{table}[h]
\centering
\caption{\textbf{Ablation on the final annealing temperatures of SCAN on \textsc{Madelon} dataset.}
Rows vary the final local temperature $\mathbb{T}_{\mathrm{end}}$ and
columns vary the final global temperature $\mathbb{T}_{g,\mathrm{end}}$.
Entries report the test AUROC (higher is better) on a single train/validation/test split.
The best result is in \textbf{bold} and the second best is \underline{underlined}.
The last row and column give the marginal AUROC, i.e., the mean over the other temperature, and the best marginal values are also in bold.}
\label{tab:ablation_temperature}
\setlength{\tabcolsep}{7pt}
\begin{tabular}{c ccccc c}
\toprule
\multirow{2}{*}{$\mathbb{T}_{\mathrm{end}}$} & \multicolumn{5}{c}{$\mathbb{T}_{g,\mathrm{end}}$} & \multirow{2}{*}{Marginal} \\
\cmidrule(lr){2-6}
 & 0.01 & 0.02 & 0.03 & 0.05 & 0.10 & \\
\midrule
0.01 & 0.854 & 0.856 & 0.833 & 0.841 & 0.838 & 0.844 \\
0.03 & 0.894 & 0.873 & 0.866 & 0.860 & 0.857 & 0.870 \\
0.10 & 0.907 & 0.890 & 0.889 & 0.863 & 0.844 & \textbf{0.879} \\
0.30 & \underline{0.911} & \textbf{0.916} & 0.869 & 0.814 & 0.784 & 0.859 \\
0.50 & 0.908 & 0.866 & 0.872 & 0.821 & 0.789 & 0.851 \\
\midrule
Marginal & \textbf{0.895} & 0.880 & 0.866 & 0.840 & 0.822 & 0.861 \\
\bottomrule
\end{tabular}
\end{table}

Table~\ref{tab:ablation_temperature} reports the test AUROC of SCAN on a single split of \textsc{Madelon} over a $5 \times 5$ grid of final local temperatures $\mathbb{T}_{\mathrm{end}}$ and final global temperatures $\mathbb{T}_{g,\mathrm{end}}$.
Unlike the grid used for the other benchmark datasets, this grid includes pairs with $\mathbb{T}_{g,\mathrm{end}} \geq \mathbb{T}_{\mathrm{end}}$.

\paragraph{Effect of $\mathbb{T}_{g,\mathrm{end}}$.}
The marginal AUROC decreases from 0.895 to 0.822 as $\mathbb{T}_{g,\mathrm{end}}$ increases from 0.01 to 0.10, and the same decreasing trend appears in most rows.
A smaller global temperature strengthens the global component-selection preference in \eqref{eq:dual_temp}.
This trend supports a small $\mathbb{T}_{g,\mathrm{end}}$ on \textsc{Madelon}, although the table alone does not determine whether the gain is due to feature selection, optimization, or reduced truncation error.

\paragraph{Effect of $\mathbb{T}_{\mathrm{end}}$.}
The marginal AUROC is highest at $\mathbb{T}_{\mathrm{end}}=0.10$ (0.879) and lowest at $\mathbb{T}_{\mathrm{end}}=0.01$ (0.844).
A very small final local temperature produces more concentrated component-selection weights, which can reduce flexibility in feature selection.
The results are consistent with an advantage from retaining some local softness, although they do not directly measure the optimization mechanism.
At $\mathbb{T}_{\mathrm{end}}=0.01$, the AUROC varies little across $\mathbb{T}_{g,\mathrm{end}}$ (0.833--0.856), suggesting that the overly concentrated local weights limit the performance regardless of the global temperature.

\paragraph{Practical recommendation.}
On this split, $\mathbb{T}_{\mathrm{end}}=0.10$ attains the highest marginal AUROC with a moderate spread across global temperatures (0.844--0.907), while $\mathbb{T}_{\mathrm{end}}=0.30$ attains the highest single AUROC but is more sensitive to the global temperature (0.784--0.916).
Since these results are based on a single split, small differences between individual entries should not be over-interpreted; we therefore focus on the trends of the marginal values.
These results support $\mathbb{T}_{g,\mathrm{end}}\leq0.02$ with $\mathbb{T}_{\mathrm{end}}\in[0.10,0.30]$ for this dataset.
Final settings are selected using validation data rather than the test scores in this table.

\subsection{Choice of univariate unit function $\mathcal{B}_{j,t}$}
\label{app:choice_basis}

To assess the choice of the univariate unit function, we replace $\mathcal{B}_{j,t}$ with either a deep neural network (DNN) or the identity function and evaluate prediction on \textsc{Calhousing}.
The DNN has three hidden layers of widths $[32,16,16]$.
Table~\ref{table:various_bbT} reports results with hyperparameters selected by grid search.
The depth-1 smooth decision tree gives lower RMSE than the two alternatives in this comparison.

\begin{table}[h]
\centering
\scriptsize
\caption{\footnotesize \textbf{Results of prediction performance of SCAN based on various $\mathcal{B}_{j,t}$.}}
\label{table:various_bbT}
\begin{tabular}{cccc}
\toprule
$\mathcal{B}_{j,t}$ & DNN & Identity & Smooth tree  \\ \midrule
RMSE $\downarrow$ (SD) & 0.556 (0.01)  & 0.756 (0.01) & $\textbf{0.471}$ (0.01)  \\ \bottomrule
\end{tabular}
\end{table}

\newpage

\section{Computational Cost of Post-hoc Identification}
\label{app:cost_identification}
We measure the computational cost of the post-hoc identification on the synthetic dataset based on $f^{(1)}$ (Table~\ref{table:def_syn}) with $p \in \{10, 50, 100\}$ input features, where $X_1, \ldots, X_{10}$ follow the distributions in Table~\ref{table:dit_syn} and $X_{11}, \ldots, X_p \overset{\text{iid}}{\sim} \mathrm{U}(-1,1)$.

For each $p$, we generate $n = 15{,}000$ observations with signal-to-noise ratio $5$ and repeat the experiment five times with different random seeds for the data generation, the data split and model training.
In each repetition, the components are identified with $\epsilon_{\alpha} = 0.01$.
The integrals in (\ref{eq:purify}) are approximated with $Q \in \{10, 100, 1000\}$ empirical quantiles of each feature, or with $m = 2000$ points drawn from the product of the empirical marginal distributions.

We report the runtime and the peak GPU memory of the post-hoc identification.
To check that the approximation does not affect component selection, we also report the AUROC of the component importance $\|\hat f^{*}_S\|_{2,n}$ for each order $|S| \in \{1, 2, 3\}$.
\begin{table}[h]
\centering
\footnotesize
\caption{\textbf{Runtime, peak GPU memory, and component selection AUROC of the post-hoc identification on $f^{(1)}$.}
$Q$: number of empirical quantiles of each feature used to approximate the integrals in (\ref{eq:purify}). $m$: number of points drawn from the product of the empirical marginals.
Mean (standard deviation) over five repetitions.}
\label{table:cost_identification}
\begin{tabular}{c l c c c c c}
\toprule
 & & & & \multicolumn{3}{c}{AUROC} \\
\cmidrule(lr){5-7}
$p$ & Integration points & Runtime (s) & Peak memory (MB) & $|S| = 1$ & $|S| = 2$ & $|S| = 3$ \\
\midrule
\multirow{4}{*}{10}
 & $Q = 10$   & 1.75 (0.55) & 13 (4)       & 0.956 (0.061) & 0.722 (0.044) & 0.588 (0.089) \\
 & $Q = 100$  & 1.76 (0.55) & 18 (5)       & 0.933 (0.061) & 0.732 (0.050) & 0.590 (0.080) \\
 & $Q = 1000$ & 1.76 (0.55) & 106 (31)     & 0.933 (0.061) & 0.733 (0.049) & 0.598 (0.075) \\
 & $m = 2000$ & 1.81 (0.59) & 243 (44)     & 0.933 (0.061) & 0.734 (0.052) & 0.600 (0.077) \\
\midrule
\multirow{4}{*}{50}
 & $Q = 10$   & 2.51 (0.23) & 416 (99)     & 0.980 (0.026) & 0.924 (0.016) & 0.939 (0.016) \\
 & $Q = 100$  & 2.57 (0.28) & 424 (87)     & 0.983 (0.022) & 0.927 (0.019) & 0.937 (0.016) \\
 & $Q = 1000$ & 4.30 (1.25) & 1,312 (67)   & 0.983 (0.022) & 0.929 (0.019) & 0.939 (0.014) \\
 & $m = 2000$ & 6.22 (2.32) & 2,316 (67)   & 0.983 (0.023) & 0.928 (0.019) & 0.938 (0.015) \\
\midrule
\multirow{4}{*}{100}
 & $Q = 10$   & 2.34 (0.26) & 224 (69)     & 0.944 (0.027) & 0.936 (0.062) & 0.926 (0.137) \\
 & $Q = 100$  & 2.35 (0.26) & 255 (59)     & 0.947 (0.029) & 0.936 (0.062) & 0.926 (0.137) \\
 & $Q = 1000$ & 2.59 (0.31) & 1,159 (59)   & 0.947 (0.029) & 0.936 (0.062) & 0.926 (0.137) \\
 & $m = 2000$ & 2.95 (0.41) & 2,164 (59)   & 0.947 (0.029) & 0.936 (0.062) & 0.926 (0.137) \\
\bottomrule
\end{tabular}
\end{table}
Table~\ref{table:cost_identification} shows that the post-hoc identification is inexpensive.
With $Q = 10$ quantiles, it takes about $2$-$3$ seconds and at most a few hundred megabytes of GPU memory for a SCAN with $T=1,000$, even for $p = 100$.
The measured cost does not increase monotonically with $p$: runtime at $p=100$ is comparable to that at $p=50$, and peak memory is smaller.
This agrees with the analysis in Section~\ref{sec:identification}: the number of required projections depends on the selected component index sets and their downward closure, rather than on all subsets of the input features. \\
$\newline$
Using more integration points makes the computation more expensive, mostly in memory.
When we increase the number of points from $10$ to $2{,}000$, the memory use becomes about $5$--$19$ times larger, but the runtime becomes at most about $2.5$ times longer.
The component selection AUROC changes little with the integration rule.
For every $p$ and order, the AUROC with $Q=10$ differs from that with $m=2000$ by at most $0.023$, less than the reported standard deviations.
Based on these results, we use 100 empirical quantiles of each feature in the post-hoc identification.

\newpage

\section{Maximum Order of the Learned Components}
\label{app:structural_order}

Since the post-hoc identification computes a component for every subset of each $S_t$, its cost is mainly determined by the orders $|S_t|$ rather than by $p$.
A desirable model should therefore learn component index sets whose downward closure contains $\mathcal{S}^{*}$ but is otherwise small.
As a baseline, we consider XGB, which builds an additive model of trees by boosting, where each tree is grown by greedily selecting the input features.
Since each tree depends only on its split variables, the fitted function can be written as $f = \sum_{S} g_S$, where $g_S$ is the sum of the trees whose set of split variables is $S$.
We compare the component index sets learned by SCAN and XGB on the synthetic data in Section~\ref{app:syn_data}.

\begin{table}[h]
\centering
\caption{\textbf{Test RMSE and maximum order $\max_t |S_t|$ on the synthetic data ($p = 50$).} Mean $\pm$ standard deviation over 5 random splits.}
\label{tab:xgb_order}
\resizebox{\textwidth}{!}{%
\begin{tabular}{lcccccc}
\toprule
 & \multicolumn{2}{c}{$f^{(1)}$} & \multicolumn{2}{c}{$f^{(2)}$} & \multicolumn{2}{c}{$f^{(3)}$} \\
\cmidrule(lr){2-3} \cmidrule(lr){4-5} \cmidrule(lr){6-7}
Model & Test RMSE & Max order & Test RMSE & Max order & Test RMSE & Max order \\
\midrule
True function & -- & 4 & -- & 4 & -- & 5 \\
\midrule
SCAN & 0.3162 $\pm$ 0.0051 & 5.0 $\pm$ 1.0 & 0.9519 $\pm$ 0.0220 & 4.6 $\pm$ 0.5 & 0.5937 $\pm$ 0.0148 & 5.8 $\pm$ 0.8 \\
XGB (\texttt{max\_depth} $= 6$) & 0.3219 $\pm$ 0.0034 & 37.2 $\pm$ 0.8 & 0.9948 $\pm$ 0.0056 & 40.0 $\pm$ 1.0 & 0.6154 $\pm$ 0.0050 & 37.6 $\pm$ 0.9 \\
XGB (\texttt{max\_depth} $= 50$) & 0.3348 $\pm$ 0.0020 & 50.0 $\pm$ 0.0 & 1.0800 $\pm$ 0.0116 & 50.0 $\pm$ 0.0 & 0.6919 $\pm$ 0.0026 & 50.0 $\pm$ 0.0 \\
\bottomrule
\end{tabular}%
}
\end{table}

Table~\ref{tab:xgb_order} shows that SCAN attains the lowest test RMSE with a much smaller maximum order of component than XGB.
The trees of XGB use 37 to 50 of the 50 inputs, so their downward closure contains almost all subsets of $[p]$.
The component index sets of SCAN are slightly larger than the true ones, but the importance scores still recover the true components with high AUROC on the same data (see Section~\ref{sec:comp_selc}).


\end{document}